\documentclass[twoside,11pt]{article}

\usepackage{amsmath}
\usepackage{amsthm}
\usepackage{bbm}
\usepackage{psfrag}
\usepackage{enumerate}
\usepackage{enumitem}
\usepackage{algorithm}
\usepackage{algpseudocode}
\usepackage{url} 
\usepackage{subfig}
\usepackage{sectsty}
\usepackage{secdot}
\usepackage{siunitx}
\usepackage{etoolbox}
\usepackage{booktabs}
\usepackage[hang,flushmargin]{footmisc}
\usepackage[preprint]{jmlr2e}
\usepackage{xcolor}
\usepackage{float}
\usepackage{multirow}

\newcommand{\simiid}{\overset{\textrm{i.i.d.}}{\sim}}
\newcommand{\simind}{\overset{\textrm{ind.}}{\sim}}
\newcommand{\toProb}{\overset{p}{\to}}
\newcommand{\toDist}{\overset{d}{\to}}
\newcommand{\EE}{\mathbb{E}}

\newcommand{\MM}{\mathcal{M}_i(\delta)}
\newcommand{\vx}[1]{v_{#1}(x)}
\newcommand{\ux}[1]{\mathbbm{1}(U_{#1} \leq z)}
\newcommand{\Var}{\text{Var}}
\newtheorem{assumption}{Assumption}
\newtheorem{theorem2}{Theorem}
\newtheorem{proposition2}{Proposition}
\newtheorem{lemma2}{Lemma}
\ifdefined\proof
    \renewenvironment{proof}[1][Proof]{\par\noindent{\bf #1\ }}{\hfill\BlackBox}
\else
    \newenvironment{proof}[1][Proof]{\par\noindent{\bf #1\ }}{\hfill\BlackBox}
\fi

\usepackage{lastpage}
\jmlrheading{22}{2021}{1-\pageref{LastPage}}{8/18; Revised 11/20}{1/21}{18-558}{Benjamin Lu and Johanna Hardin}

\ShortHeadings{Random Forest Prediction Error Estimation}{Lu and Hardin}
\firstpageno{1}

\begin{document}

\title{A Unified Framework for Random Forest Prediction Error Estimation}

\author{\name Benjamin Lu \email b.lu@berkeley.edu \\
       \addr Department of Statistics\\
       University of California, Berkeley\\
       Evans Hall\\
       Berkeley, CA 94720, USA
       \AND
       \name Johanna Hardin \email jo.hardin@pomona.edu \\
       \addr Department of Mathematics\\
       Pomona College\\
       610 N. College Ave.\\
       Claremont, CA 91711, USA}

\editor{Boaz Nadler}

\maketitle

\begin{abstract}
We introduce a unified framework for random forest prediction error estimation based on a novel estimator of the conditional prediction error distribution function. Our framework enables simple plug-in estimation of key prediction uncertainty metrics, including conditional mean squared prediction errors, conditional biases, and conditional quantiles, for random forests and many variants. Our approach is especially well-adapted for prediction interval estimation; we show via simulations that our proposed prediction intervals are competitive with, and in some settings outperform, existing methods. To establish theoretical grounding for our framework, we prove pointwise uniform consistency of a more stringent version of our estimator of the conditional prediction error distribution function. The estimators introduced here are implemented in the {\ttfamily R} package {\ttfamily forestError}.
\end{abstract}


\section{Introduction}
\label{sec:intro}
Random forests and other tree-based methods are often used for regression---that is, to relate a real-valued response $Y$ to covariates $X$ \citep{criminisi2010, grimm2008, wei2010}. The objective in many of these applications is to predict the unknown responses of observations given their covariates. We denote such a prediction by $\hat{\varphi}(X)$. For example, researchers in precision medicine seek to predict the health outcomes of individual patients under some treatment regime given patient, clinical, and environmental characteristics, with the ultimate goal of developing individualized therapies for patients \citep{fang2018}. Other researchers seek to predict bird migration patterns to reduce collisions with airplanes, wind turbines, and buildings \citep{vanDoren2018}.

When using any regression method for prediction, quantifying the uncertainty associated with the predictions can enhance their practical value. One central function for quantifying uncertainty is the conditional prediction error distribution, which, letting $E := Y - \hat{\varphi}(X)$ denote the error of a prediction, is given by
\[
F_E\left(e \mid x\right) := \Pr\left(E \leq e \mid X = x\right) = \Pr\left(Y - \hat{\varphi}(X) \leq e \mid X = x\right).
\]
The conditional prediction error distribution can be mapped to a number of useful parameters that characterize prediction uncertainty. For example, the conditional mean squared prediction error,
\[
\text{MSPE}(x) := \mathbb{E}\left[(Y - \hat{\varphi}(X))^2 \mid X = x\right] = \int e^2f_E(e \mid x)\,de,
\]
can summarize how erroneous a given point prediction is expected to be. Additionally, the conditional bias,
\[
\text{Bias}(x) := \mathbb{E}\left[\hat{\varphi}(X) - Y \mid X = x\right] = -\int ef_E(e \mid x)\,de,
\]
measures systematic over- or under-prediction of the responses of units with a given set of covariates. 
Finally, the $\alpha$-quantiles of the conditional prediction error distribution,
\[
Q^\alpha_E(x) := \inf\left\{e:F_E\left(e \mid x\right) \geq \alpha\right\},
\]
can be used to construct a conditional prediction interval containing the unknown response of a given observation with a specified probability.

This paper proposes a method of estimating the conditional prediction error distribution $F_E\left(e \mid x\right)$ of random forests. With this estimate, conditional mean squared prediction error, conditional bias, conditional quantiles, and other parameters of the distribution can all be estimated with ease. By contrast, the current literature on characterizing the uncertainty of random forest predictions has been piecemeal. For example, existing estimators of conditional biases and conditional response quantiles were developed separately, rely on different assumptions, and are computed by separate algorithms. Thus, the central contribution of this paper is a unified framework for assessing random forest prediction uncertainty, with a suite of estimators that empirically are competitive with, and in some cases outperform, existing methods, particularly for the tasks of prediction interval estimation and quantile regression.

In addition to creating a unified framework, our method is general in the sense that it can be implemented for many variants of the random forest algorithm. For example, it is compatible with a wide range of decision tree algorithms that partition the covariate space based on different criteria, such as generalized random forests \citep{Athey2019}, as well as various resampling and subsampling regimes that have been examined in recent literature \citep{biau2008}. It can also be naturally adapted to augmentations of the random forest algorithm, such as local linear forests \citep{Friedberg2019}.

The remainder of this manuscript is organized as follows. Section \ref{sec:review} reviews the literature on estimating parameters of $F_E(e \mid x)$ that are commonly of interest. We establish the setting and relevant notation for our problem in Section \ref{sec:setup}. Then, we introduce in Section \ref{sec:prederror} our proposed estimator of $F_E(e \mid x)$ and show how it enables simple plug-in estimation of parameters of $F_E(e \mid x)$. In Section \ref{sec:sim}, we assess the empirical performance of some of these resulting plug-in estimators. In Section \ref{sec:theory}, we propose and prove uniform consistency of an estimator of $F_E(e \mid x)$ that is similar to but more stringently constructed than the one proposed in Section \ref{sec:prederror}. Section \ref{sec:conclusion} concludes.

\section{Related Work}
\label{sec:review}

To our knowledge, we are the first to propose a method of estimating the conditional prediction error distribution of random forests. However, random forest mean squared prediction error, bias, and prediction intervals have each been studied individually in previous works. We briefly review the literature on each in turn.

The most, and perhaps only, widely used summary metric for random forest prediction error is the unconditional mean squared prediction error,
\[
\text{MSPE} := \mathbb{E}\left[(Y - \hat{\varphi}(X))^2\right],
\]
which is usually estimated by an out-of-bag procedure \citep{breiman1996, liaw2002}. We propose an estimator of the conditional mean squared prediction error $\text{MSPE}(x)$, which, as we illustrate in Section \ref{sec:extensions}, is often a more informative metric.

The literature on random forest bias has been more active. \citet{Wager2018} show that random forests are biased and provide a bound on the magnitude of the bias under certain assumptions about the tree-growing mechanism and the underlying data distribution. \citet{Ghosal2018} leverage this work to investigate a method of bias correction, initially proposed by \citet{breiman1999bag}, that fits a random forest on the out-of-bag prediction errors to directly model the bias. This boosting approach, which is similar to gradient boosting \citep{friedman2001}, is also studied by \citet{zhang2012}, who propose additional model-based bias corrections for random forests. \citet{hooker2018} propose a different method of bias correction that approximates the classic bootstrap bias estimation procedure \citep{efron1994} in a more computationally efficient way. We contribute to this literature by proposing a new bias correction procedure and comparing it to the boosting method examined by \citet{Ghosal2018} and \citet{zhang2012}.

The literature on prediction interval estimation for random forests began with the development by \citet{meinshausen2006} of quantile regression forests, a random forest-based algorithm that enables consistent estimation of conditional prediction intervals. Since then, \citet{Athey2019} have proposed generalized random forests, a method of estimating quantities identified by local moment conditions that grows trees specifically designed to express heterogeneity in the quantity of interest. They show that their algorithm can be used for quantile regression. Additionally, \citet{zhang2019} propose estimating prediction intervals using the empirical quantiles of a random forest's out-of-bag prediction errors. More broadly, conformal inference offers a generic way of estimating prediction intervals that can be applied to virtually any estimator of the regression function, including random forests \citep{Lei2014, Lei2018, Johansson2014}. We add to this literature by proposing a new prediction interval estimator and assessing the strengths and weaknesses of each method through simulation.

Finally, we distinguish our work from two segments of the random forest literature. First, although the prediction interval estimator we propose in Section \ref{sec:extensions} superficially resembles the proposal by \citet{zhang2019} mentioned above, this paper differs from their work in three major respects. First, as an overarching matter, \citet{zhang2019} focus solely on prediction interval estimation for random forests built via the classification and regression tree (CART) algorithm. By contrast, our work establishes a suite of easily computed prediction uncertainty metrics, of which prediction intervals are just one, for a broad class of tree-based algorithms. Second, their prediction intervals provide only unconditional coverage at the desired rate and by construction have the same width for all test observations. By contrast, our prediction intervals provide conditional coverage at the desired rate and adapt to changes in the shape of the conditional response distribution across the covariate space. We illustrate this distinction, which is often vital in realistic applications, via simulation in Section \ref{sec:sim}. Third, the mathematical justifications for the asymptotic properties of their estimator rely on assumptions different from the ones we use in Section \ref{sec:theory}.

Second, we emphasize that our work is largely separate from the literature on the use of random forest-based algorithms for conditional mean estimation and inference \citep{sexton2009, wager2014, Mentch2016, Wager2018}. Although, for many regression methods, the point estimator for an individual response is equivalent to the point estimator for the conditional mean, the statistical challenges of conditional mean estimation are different from those of prediction error estimation. For example, many methods of conditional mean estimation and inference invoke some type of central limit theorem to characterize their estimators' behavior; such approaches are generally less applicable to prediction error estimation, which concerns individual responses rather than their expected value.

\section{Setup and Notation}
\label{sec:setup}
Consider an observed training sample $\mathcal{D}_n := \{(X_i, Y_i)\}_{i = 1}^{n}$, where $(X_i, Y_i) \simiid \mathbb{P}$ for some distribution $\mathbb{P}$, $X_i \in \mathcal{X}$ is a $p$-dimensional covariate with support $\mathcal{X}$, and $Y_i \in \mathbb{R}$ is a real-valued response with a continuous conditional distribution function $F_Y(y \mid x)$. For convenience, we let $Z_i := (X_i, Y_i)$. A standard implementation of random forests fits a tree on each of $B$ bootstrap samples of the training set $\mathcal{D}_{n, 1}^*, \ldots, \mathcal{D}_{n, B}^*$ using some algorithm, such as the CART algorithm, with the $b^{\text{th}}$ tree's construction governed by a random parameter $\theta_b$ drawn i.i.d. from some distribution independently of $\mathcal{D}_n$ \citep{breiman2001}. Included in $\theta_b$, for example, might be the randomization of eligible covariates for each split. Each tree splits its bootstrap training sample $\mathcal{D}_{n, b}^*$ into terminal nodes; each split corresponds to a partitioning of the predictor space $\mathcal{X}$ into rectangular subspaces. For the $b^{\text{th}}$ tree, let $\ell(x, \theta_b)$ index the terminal node corresponding to the subspace containing $x$, and let $R_{\ell(x, \theta_b)}$ denote the subspace itself. With this notation, we introduce the following terminology; to our knowledge, the literature has not settled on a term for observations satisfying Definition \ref{def:cohab}, although the underlying concept is closely related to the notion of ``connection functions'' in the characterization by \citet{scornet2016} of random forests as kernel methods.

\begin{definition}
\label{def:cohab}
A training observation $Z_i = (X_i, Y_i)$ is a \textbf{cohabitant} of $x$ in tree $b$ if and only if $\ell(X_i, \theta_b) = \ell(x, \theta_b)$.
\end{definition}

When predicting the response of a test observation with realized covariate value $x$, each tree in the random forest employs a weighted average of the in-bag training responses, with weights corresponding to cohabitation. In particular, the in-bag weight given to the $i^{\text{th}}$ observation in the $b^{\text{th}}$ tree is
\[
w_{i}(x, \theta_b) := \frac{\#\{Z_i \in \mathcal{D}^{*}_{n,b}\}\mathbbm{1}(X_i \in R_{\ell(x, \theta_b)})}{\sum_{j = 1}^{n}\#\{Z_j \in \mathcal{D}^{*}_{n,b}\}\mathbbm{1}(X_j \in R_{\ell(x, \theta_b)})},
\]
where $\#\{Z_i \in \mathcal{D}^*_{n,b}\}$ denotes the number of times the $i^{\text{th}}$ observation is in $\mathcal{D}^*_{n,b}$ and $\mathbbm{1}(\cdot)$ is the indicator function. Note that $w_i(x, \theta_b)$ is a random variable based in part on $X_i$, hence the subscript $i$. The random forest prediction of the response of units with covariate value $x$ is the average of the tree predictions:
\[
\hat{\varphi}(x) := \frac{1}{B}\sum_{b = 1}^{B}\sum_{i = 1}^{n}w_i(x, \theta_b)Y_i.
\]

It is well-known that, with a sufficiently large number of trees grown on bootstrap samples of $\mathcal{D}_n$, each training observation will be out of bag---that is, not included in the bootstrap sample---for any number fewer than $\left \lfloor{B(1 - 1 / n)^n}\right \rfloor$ of the trees with high probability. Thus, we can define out-of-bag analogues to $w_i(x, \theta_b)$ and $\hat{\varphi}(x)$. The out-of-bag weight given to the $i^{\text{th}}$ training observation is the proportion of times the $i^{\text{th}}$ training observation is an out-of-bag cohabitant of $x$, relative to all training observations:
\[
v_i(x) := \frac{\sum_{b = 1}^{B}\mathbbm{1}(Z_i \notin \mathcal{D}^*_{n,b}\text{ and }X_i \in R_{\ell(x,\theta_b)})}{\sum_{j = 1}^{n}\sum_{b = 1}^{B}\mathbbm{1}(Z_j \notin \mathcal{D}^*_{n,b}\text{ and }X_j \in R_{\ell(x, \theta_b)})}.
\]
This is a random variable based in part on $X_i$, hence the subscript $i$. Notice that, unlike $w_i(x, \theta_b)$, $v_i(x)$ is defined over all trees because $x$ is guaranteed an in-bag cohabitant in each tree but not an out-of-bag cohabitant. The out-of-bag prediction of the $i^{\text{th}}$ training unit is the average prediction of the unit's response among the trees for which the unit is out of bag:
\[
\hat{\varphi}^{(i)}(X_i) := \frac{1}{\sum_{b = 1}^{B}\mathbbm{1}(Z_i \notin \mathcal{D}^*_{n,b})}\sum_{b:Z_i \notin \mathcal{D}^*_{n,b}}\sum_{j = 1}^{n}w_j(X_i, \theta_b)Y_j.
\]

\section{A Unified Framework for Assessing Prediction Uncertainty}
\label{sec:prederror}

In this section, we present a practical implementation of our proposed method of estimating the conditional prediction error distribution $F_E(e \mid x)$ and show how it facilitates estimation of conditional mean squared prediction errors, conditional biases, and conditional prediction intervals. This practical implementation is similar in spirit to but less stringent in its construction than our more rigorous method of estimating $F_E(e \mid x)$, which we detail and prove is uniformly consistent in Section \ref{sec:theory}. Nonetheless, we present the practical version first to build intuition, demonstrate its viability in empirical applications (see Section \ref{sec:sim}), and suggest potential areas of future research. Similar simplifications have been made in other recent work on random forests \citep{meinshausen2006}. The estimators discussed in this section are implemented in the {\ttfamily{R}} package {\ttfamily{forestError}}.

\subsection{Estimating the Conditional Prediction Error Distribution}

The practical implementation of our proposed method estimates $F_E(e \mid x)$ by out-of-bag weighting of the out-of-bag prediction errors:
\begin{equation}
\label{eq:mainresult}
\hat{F}_E(e \mid x) := \sum_{i = 1}^{n}v_i(x)\mathbbm{1}\left(Y_i - \hat{\varphi}^{(i)}(X_i) \leq e\right).
\end{equation}
This approach is grounded in the principle that, because training observations are not used in the construction of trees for which they are out of bag, the relationship between a training observation and the subset of trees for which it is out of bag is analogous to the relationship between the test observation and the random forest when the number of training observations and trees is large. In particular, not only are the out-of-bag prediction errors a reasonable proxy for the error of future test predictions \textit{in general}, but also the out-of-bag prediction errors of training observations that are more frequently out-of-bag cohabitants of a given test observation make better proxies for the prediction error of that \textit{specific} test observation than the out-of-bag prediction errors of training observations that are out-of-bag cohabitants less often. Broadly speaking, this notion of similarity that motivates our use of the out-of-bag weights $v_i(x)$ also underpins the ``proximity'' measure commonly included in random forest implementations. But there are slight differences between these two similarity measures. In particular, proximity is traditionally measured between pairs of training observations, counts in-bag cohabitation as well as out-of-bag cohabitation, and is normalized by the number of trees in the forest \citep{liaw2002, breiman2002manual}. By contrast, $v_i(x)$ is measured between the training observations and a test point of interest, counts only out-of-bag cohabitation, and is normalized to sum to one.

One minor caveat for the analogy between out-of-bag training observations and test observations is that fewer trees are used to generate out-of-bag predictions. In this respect, the out-of-bag errors more closely resemble test errors from a fraction of the random forest's trees, chosen randomly. However, as the following proposition shows, the distribution of prediction errors $E^*$ from a non-zero fraction of the $B$ trees in a random forest becomes arbitrarily similar to the distribution of prediction errors $E$ from the full random forest as $B$ increases. The proof is provided in Appendix \ref{ap:proof}.
\begin{proposition2}
\label{prop:errorConverge}
For every $x \in \mathcal{X}$,
\[
\lim_{B \to \infty}\sup_{e \in \mathbb{R}}\left|F_{E^*}(e \mid x) - F_{E}(e \mid x)\right| = 0.
\]
\end{proposition2}
Other issues, primarily concerning the dependence relations induced by the construction of the random forest and $\hat{F}_E\left(e \mid x\right)$, prevent us from proving that $\hat{F}_E(e \mid x)$ is consistent. These issues, which touch on recent areas of research, are discussed further in Section \ref{sec:theory}, where we prove uniform consistency of a similar but more stringently constructed estimator of $F_E(e \mid x)$ (Theorem \ref{th:main}). However, we believe, based on simulations presented in Section \ref{sec:sim} and Appendix \ref{ap:stringent}, that these issues are minor in practice and that $\hat{F}_{E}(e \mid x)$ as defined in this section empirically performs as well as our more stringently constructed estimator.

\subsection{Extensions}
\label{sec:extensions}
Estimators for conditional mean squared prediction errors, conditional biases, and conditional prediction intervals follow immediately by plugging in $\hat{F}_E(e \mid x)$. We describe each in turn.

\subsubsection*{Conditional Mean Squared Prediction Error}
We propose a plug-in estimator for the conditional mean squared prediction error $\text{MSPE}(x)$ that averages the squared out-of-bag prediction errors over $\hat{F}_E(e \mid x)$:
\[
\widehat{\text{MSPE}}(x) := \int e^2\hat{f}_E(e \mid x)\, de = \sum_{i = 1}^{n}v_i(x)\left(Y_i - \hat{\varphi}^{(i)}(X_i)\right)^2.
\]
To our knowledge, no other method of estimating $\text{MSPE}(x)$ has been proposed. Current implementations of random forests, such as the {\ttfamily R} package {\ttfamily randomForest} \citep{liaw2002}, instead generally estimate $\text{MSPE}$ by the unweighted average of the squared out-of-bag prediction errors; we denote this estimator by $\widehat{\text{MSPE}}$. While $\text{MSPE}$ can be an informative summary of the predictive performance of the random forest overall, $\text{MSPE}(x)$ is usually more appropriate for assessing the reliability of any individual prediction.

\begin{figure}
    \centering
    \includegraphics[width = 0.496\linewidth]{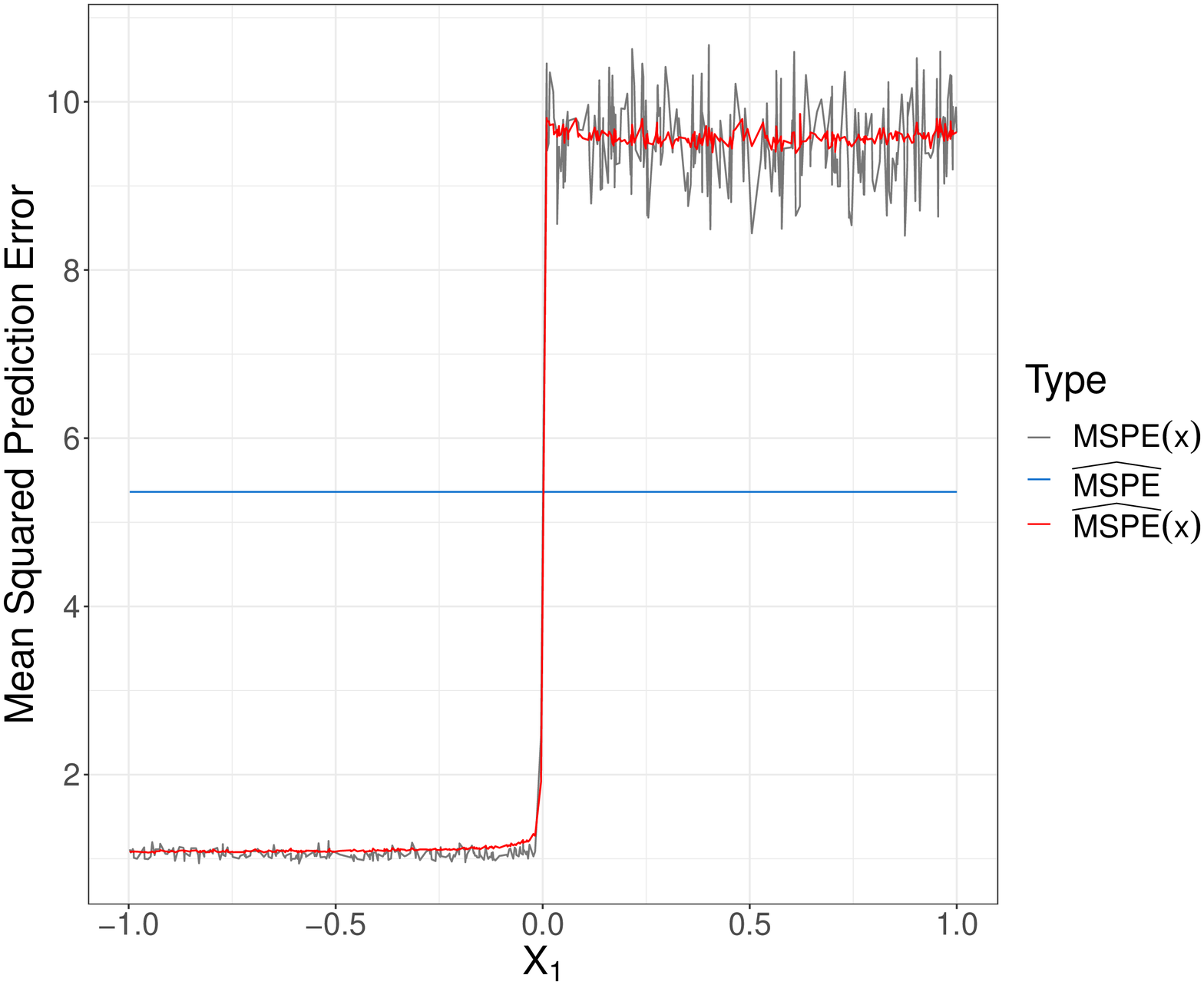}
    \caption{Comparison of $\widehat{\text{MSPE}}$ and $\widehat{\text{MSPE}}(x)$ behavior. The data were simulated as $X \simiid \text{Unif}[-1, 1]^{10}$ and $Y \simind \mathcal{N}\left(10\cdot\mathbbm{1}\left(X_{1} > 0\right), (1 + 2\cdot\mathbbm{1}(X_{1} > 0))^2\right)$.}
    \label{fig:mspemotive}
\end{figure}

Figure \ref{fig:mspemotive} illustrates this distinction between unconditional and conditional mean squared prediction error. To create this figure, we repeatedly drew 1,000 training observations $X \simiid \text{Unif}[-1, 1]^{10}$ with response $Y \simind \mathcal{N}\left(10\cdot\mathbbm{1}(X_{1} > 0), (1 + 2\cdot\mathbbm{1}(X_{1} > 0))^2\right)$. Note that, throughout this paper, we drop the subscript $i$ when discussing simulations for notational simplicity. For each draw, we fit a random forest to the training observations and predicted 500 test observations whose covariate values were fixed across the simulation repetitions but whose response values were randomly sampled from the same distribution as the training data. Figure \ref{fig:mspemotive} plots the average $\widehat{\text{MSPE}}$, the average $\widehat{\text{MSPE}}(x)$ of each test point, and the actual $\text{MSPE}(x)$ of each test point against $X_{1}$. As expected, $\text{MSPE}(x)$ is larger for test observations with $X_1 > 0$. Our estimator $\widehat{\text{MSPE}}(x)$ reflects this difference in prediction uncertainty, whereas $\widehat{\text{MSPE}}$, while descriptive of global prediction error, does not accurately assess the error one would expect from any individual prediction.

\subsubsection*{Conditional Bias}
Our proposed plug-in estimator for the conditional bias is the average of the out-of-bag prediction errors over $\hat{F}_E(e \mid x)$:
\[
\widehat{\text{Bias}}(x) := -\int e\hat{f}_E(e \mid x)\, de = \sum_{i = 1}^{n}v_i(x)\left(\hat{\varphi}^{(i)}(X_i) - Y_i\right).
\]
Thus, our bias-corrected random forest prediction at $x$ is given by
\[
\hat{\varphi}^{\text{BC}}(x) := \hat{\varphi}(x) - \widehat{\text{Bias}}(x).
\]
We compare the empirical performance of $\hat{\varphi}^{\text{BC}}(x)$ to that of the boosting method investigated by \citet{zhang2012} and \citet{Ghosal2018} in Section \ref{sec:bias_sim}.

\subsubsection*{Conditional Prediction Intervals and Response Quantiles}
For a given type-I error rate $\alpha \in (0, 1)$, a conditional $\alpha$-level prediction interval $\text{PI}_\alpha(x)$ for the response at $x$ satisfies the inequality
\[
\Pr\left(Y \in \text{PI}_\alpha(X) \mid X = x\right) \geq 1 - \alpha.
\]
We propose estimating a conditional prediction interval $\text{PI}_\alpha(x)$ by adding the $\alpha / 2$ and $1 - \alpha / 2$ quantiles of $\hat{F}_E(e \mid x)$ to the random forest prediction at $x$:
\[
\widehat{\text{PI}}_\alpha(x) := \left[\hat{\varphi}(x) + \hat{Q}^{\alpha / 2}_E(x), \hat{\varphi}(x) + \hat{Q}^{1 - \alpha / 2}_E(x)\right],
\]
where $\hat{Q}^{\alpha}_E(x) := \inf\{e:\hat{F}_E(e \mid x) \geq \alpha\}$. The bounds of $\widehat{\text{PI}}_\alpha(x)$ correspond to plug-in estimates of the $\alpha / 2$ and $1 - \alpha / 2$ quantiles of the conditional response distribution at $x$. So, more generally, we propose estimating the $\alpha$-quantile of the conditional response distribution
\[
Q^\alpha_Y(x) := \inf\{y:F_Y(y \mid x) \geq \alpha\}
\]
by the plug-in estimator
\[
\hat{Q}^\alpha_Y(x) := \hat{\varphi}(x) + \hat{Q}^\alpha_E(x).
\]
We compare the empirical performance of $\widehat{\text{PI}}_\alpha(x)$ to the performance of prediction intervals obtained by other recently proposed methods in Section \ref{sec:interval_sim}.

\subsubsection*{Conditional Misclassification Rate for Categorical Outcomes}
While this paper focuses on settings in which the response is continuous, our framework extends to random forest classification of categorical outcomes as well. In this setting, one common measure of predictive accuracy is the misclassification rate $\text{MCR} := \Pr(\hat{\varphi}(X) \neq Y)$. This is commonly estimated by the unweighted out-of-bag misclassification rate of the training sample:
\[
\widehat{\text{MCR}} := \frac{1}{n}\sum_{i = 1}^{n}\mathbbm{1}(\hat{\varphi}^{(i)}(X_i) \neq Y_i)
\]
\citep{breiman2001}. By analogy to our earlier discussion of $\text{MSPE}(x)$, the conditional misclassification rate $\text{MCR}(x) := \Pr(\hat{\varphi}(X) \neq Y \mid X = x)$ is often a more informative metric, but, to our knowledge, no estimator for it has been introduced in the literature. We propose estimating $\text{MCR}(x)$ by
\[
\widehat{\text{MCR}}(x) := \sum_{i = 1}^{n}v_i(x)\mathbbm{1}(\hat{\varphi}^{(i)}(X_i) \neq Y_i).
\]
However, a detailed examination of this estimator is beyond the scope of this paper.

\section{Simulation Studies}
\label{sec:sim}
In this section, we empirically compare our proposed bias correction and prediction intervals to existing methods reviewed in Section \ref{sec:review} across a variety of synthetic and benchmark datasets. Except where otherwise specified, we applied our methods to forests grown using the CART algorithm as implemented by the {\ttfamily randomForest} package in {\ttfamily R}. Implementation details and additional results are in Appendix \ref{ap:sim}. As noted previously, we omit the subscript $i$ when discussing simulations for notational simplicity.

\subsection{Conditional Bias Estimation}
\label{sec:bias_sim}
We compare our bias-corrected random forest $\hat{\varphi}^{\text{BC}}(x)$ to the bias-corrected random forest obtained by the boosting approach examined by \citet{zhang2012}, who refer to it as ``BC3,'' and \citet{Ghosal2018}. One metric for comparison is the mean squared bias,
\[
\text{MSB} := \mathbb{E}\left[\text{Bias}(X)^2\right],
\]
where the outer expectation is taken over the distribution of covariates. A lower value of $\text{MSB}$ indicates a lower level of bias overall. Since correcting the bias of a prediction may increase the prediction variance, a second metric for comparison is the mean squared prediction error $\text{MSPE}$, which measures overall predictive accuracy accounting for both bias and variance.

We tested each method on five synthetic datasets in which the conditional means are known by design. In each dataset, the covariates were sampled as $X \simiid \text{Unif}[0, 1]^{10}$. The responses were sampled as follows.
\begin{description}
\item[Baseline:] $Y \simiid \mathcal{N}(0, 1)$.
\item[Linear:] $Y \simind \mathcal{N}(X_1, 1)$.
\item[Step:] $Y \simind \mathcal{N}\left(10\cdot\mathbbm{1}(X_1 > 1/2), 1\right)$.
\item[Exponential:] $Y = \exp\{X_1\epsilon\},$ where $\epsilon \simiid \mathcal{N}(0, 1)$.
\item[Friedman:] $Y \simind \mathcal{N}\left(10\sin(\pi X_1X_2) + 20\left(X_3 - 1/2\right)^2 + 10X_4 + 5X_5, 1\right)$ \citep{friedman1991}.
\end{description}
In each repetition of the synthetic-dataset simulations, we drew 200 training units, as \citet{zhang2012} do in their simulations, and 2,000 test units. We fit an uncorrected random forest and each bias-corrected random forest to the training set, and then predicted the responses of the sampled test units using each estimator. We then averaged the squared prediction errors. Doing this repeatedly allowed us to estimate $\text{MSPE}$. In each repetition, we also predicted the responses of a held-out set of 2,000 units whose covariate values were fixed over all repetitions. Averaging these predictions over the repetitions enabled us to estimate the mean prediction of the uncorrected random forest and each bias-corrected random forest at each of the fixed 2,000 points; we then combined this with the true conditional mean at each point, which we knew by design, to estimate $\text{MSB}$. We ran 1,000 repetitions for each synthetic dataset. We also assessed the MSPE of each estimator on the Boston Housing, Abalone, and Servo benchmark datasets via the above procedure, using the same train-test ratios as the simulations in \citet{zhang2012}. These datasets were obtained through the UCI Machine Learning Repository and the {\ttfamily{MASS}} and {\ttfamily{mlbench}} {\ttfamily{R}} packages \citep{UCIRepository, mlbenchR, MASSR}.

\begin{figure}[h]
    \centering
    \includegraphics[width = 0.496\linewidth]{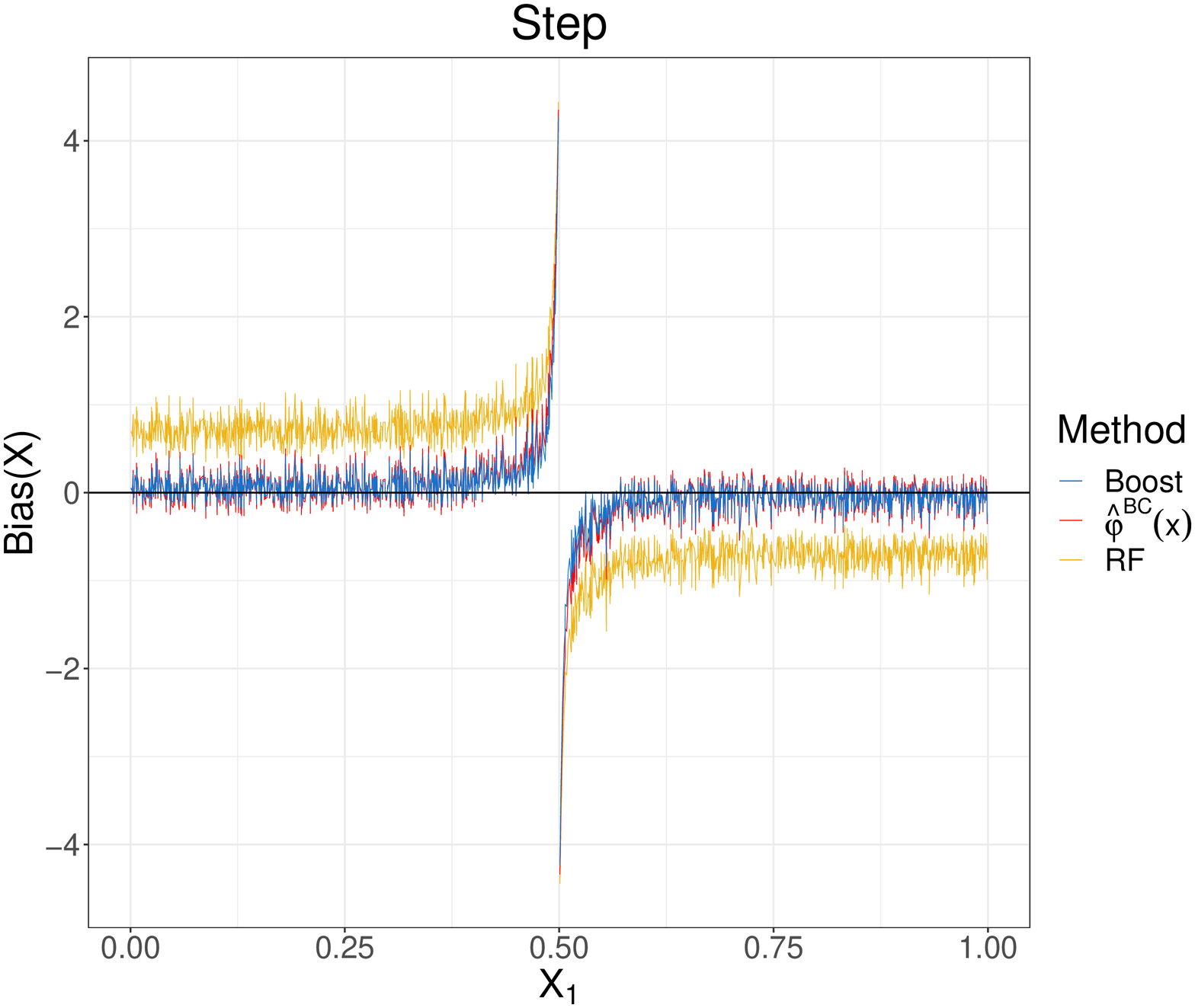}
    \includegraphics[width = 0.496\linewidth]{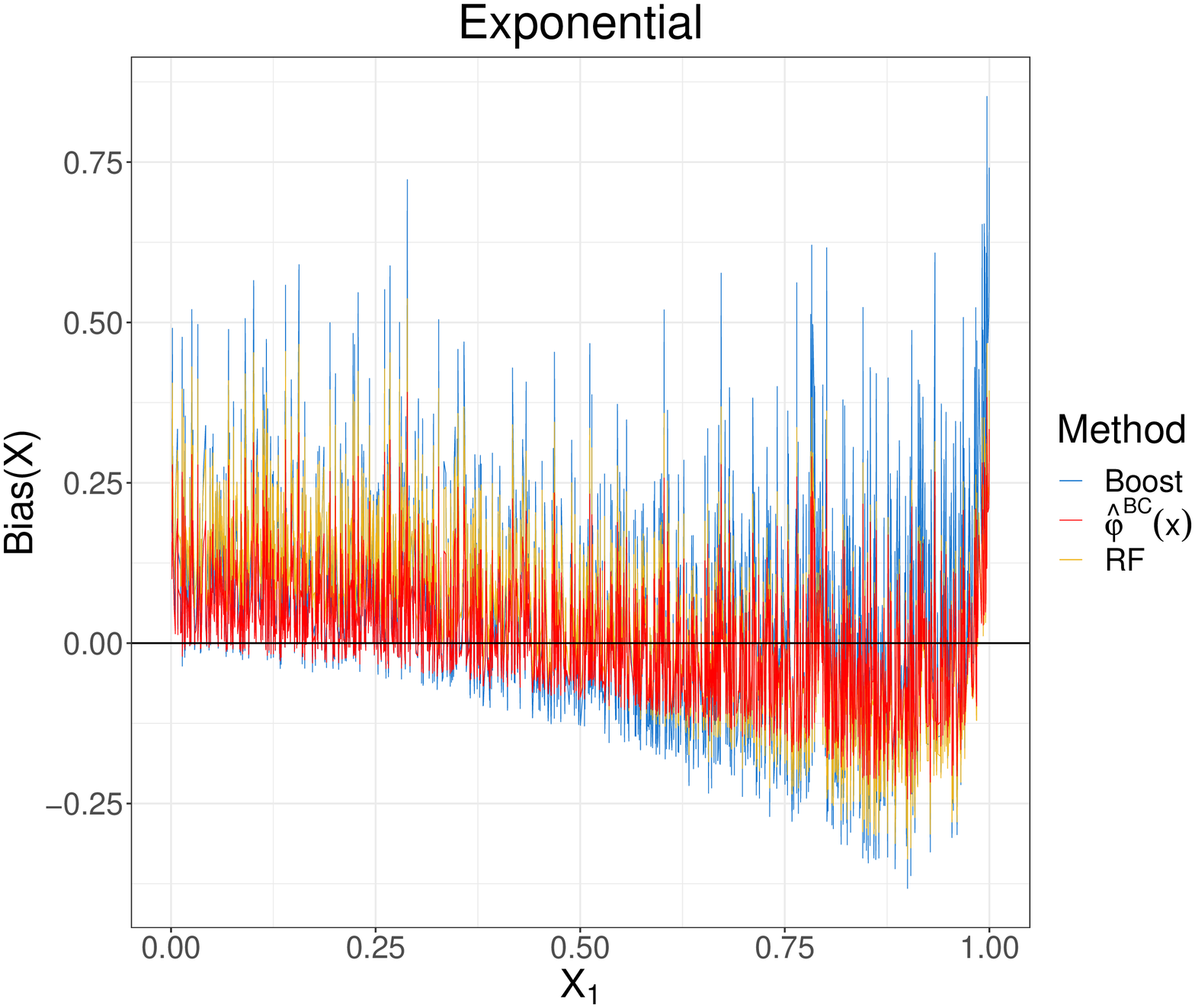}
    \includegraphics[width = 0.496\linewidth]{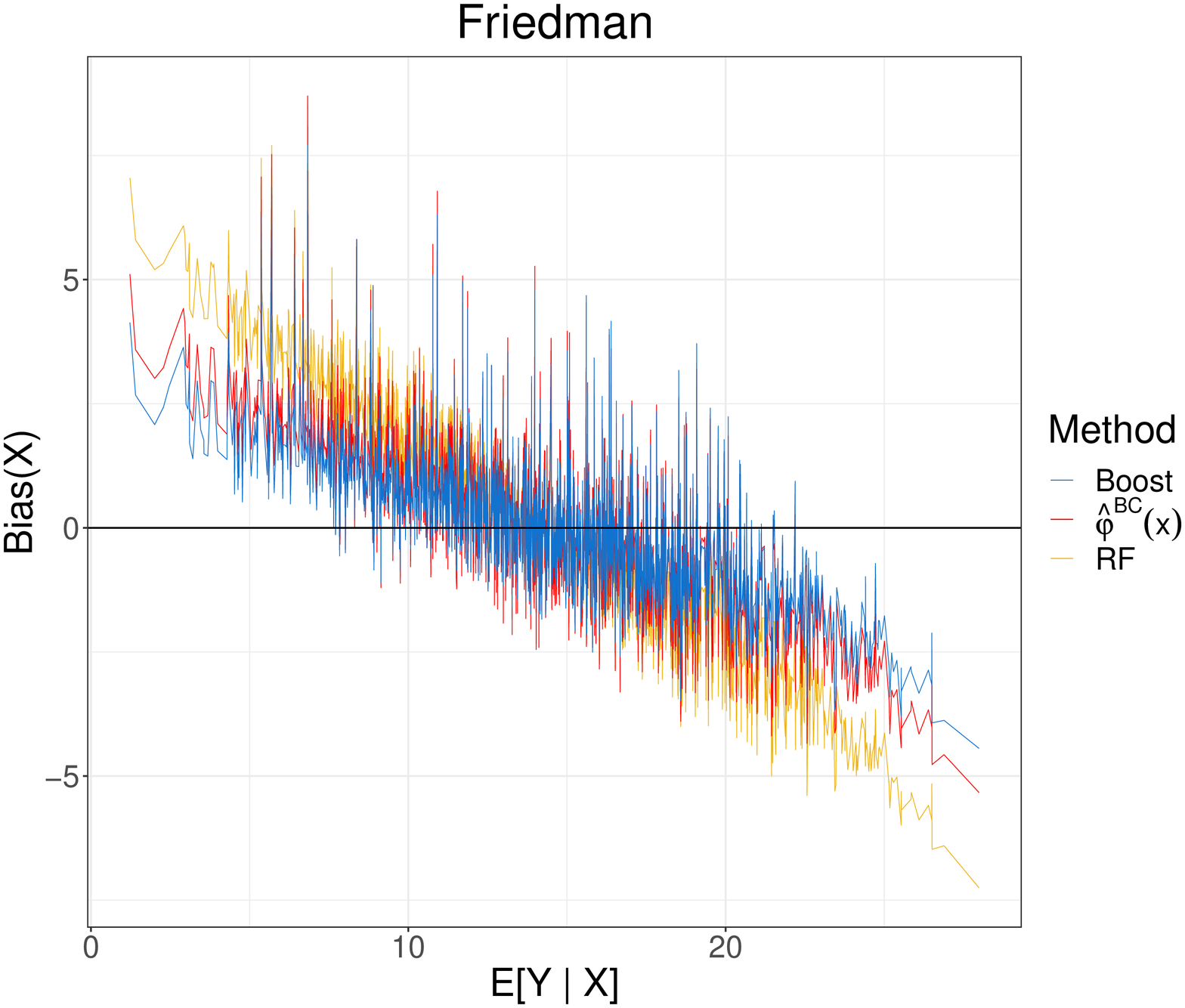}
    \includegraphics[width = 0.496\linewidth]{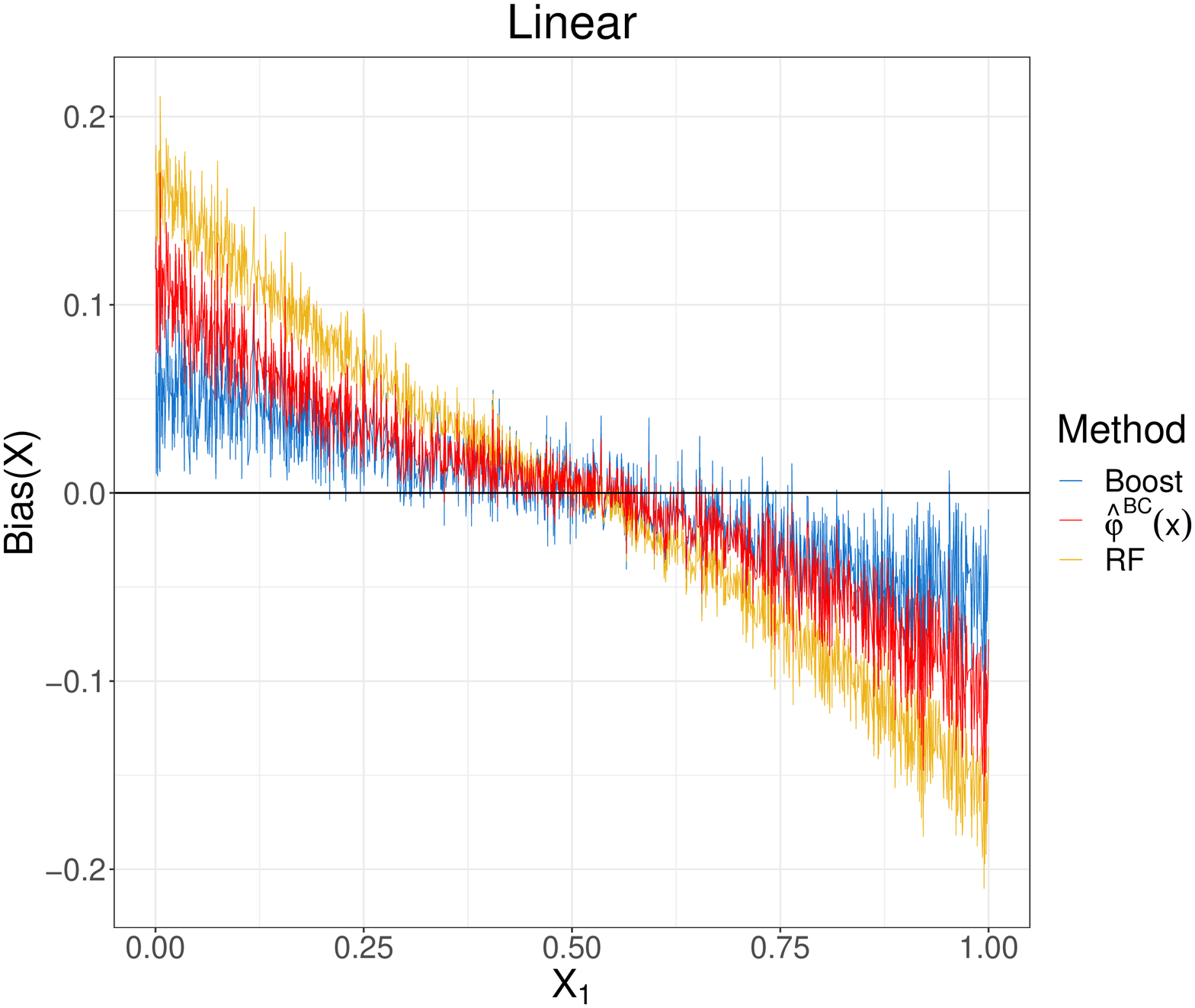}
    \caption{Conditional biases of $\hat{\varphi}^{\text{BC}}(x)$, the bias-corrected random forest based on boosting, and the uncorrected random forest for the Step, Exponential, Linear, and Friedman datasets (clockwise from top left) over 1,000 simulation repetitions.}
    \label{fig:bias_behavior}
\end{figure}

Table \ref{tab:bias_results} reports the results, and Figure \ref{fig:bias_behavior} plots the conditional bias of each method against the signaling covariate(s). Overall, our bias-corrected estimator $\hat{\varphi}^{\text{BC}}(x)$ appears to be more conservative but also more robust than the boosting approach. With respect to both MSB and MSPE, our bias correction generally improved upon but, at a minimum, did not much worse than the uncorrected random forest. By comparison, the boosting approach sometimes improved bias more than $\hat{\varphi}^{\text{BC}}(x)$ did, but in other instances it had worse bias than even the uncorrected random forest. Moreover, it sometimes reduced bias at the expense of greater variance less efficiently than $\hat{\varphi}^{\text{BC}}(x)$, as reflected in the MSPE.

\begin{table}
    \centering
    \begin{tabular}{ccccccc}
         \toprule
         & \multicolumn{3}{c}{MSB} & \multicolumn{3}{c}{MSPE} \\
         \cmidrule(r){2-4} \cmidrule(l){5-7}
         Dataset & RF & Boost & $\hat{\varphi}^{\text{BC}}(x)$ & RF & Boost & $\hat{\varphi}^{\text{BC}}(x)$\\
         \cline{1-7}
         Baseline & 0.000 & 0.000 & 0.000 & 1.063 & 1.170 & 1.085\\ 
         Linear & 0.008 & 0.002 & 0.003 & 1.074 & 1.172 & 1.095\\ 
         Step & 0.814 & 0.179 & 0.222 & 2.014 & 1.508 & 1.457\\ 
         Exponential & 0.021 & 0.031 & 0.009 & 0.997 & 1.124 & 1.002\\ 
         Friedman & 5.143 & 1.862 & 2.765 & 7.018 & 3.958 & 4.927\\ 
         Boston & - & - & - & 8.001 & 8.259 & 6.973\\ 
         Abalone & - & - & - & 4.794 & 5.031 & 4.831\\ 
         Servo & - & - & - & 26.494 & 11.779 & 17.601\\ 
         \bottomrule
    \end{tabular}
    \caption{Mean squared bias and mean squared prediction error of the uncorrected random forest, the bias-corrected random forest based on boosting, and our bias-corrected random forest $\hat{\varphi}^{\text{BC}}(x)$ for each dataset.}
    \label{tab:bias_results}
\end{table}

The greater robustness of $\hat{\varphi}^{\text{BC}}(x)$ is even more apparent in noisier settings. For example, we ran a modified version of the synthetic-dataset simulations where we set the variance of the response variable in the Baseline, Linear, Step, and Friedman data-generating processes to 100 and the variance of the noise term in the Exponential data-generating process to 4. Table \ref{tab:additional_bias_results} reports the results. Notably, the performance of $\hat{\varphi}^{\text{BC}}(x)$ does not deteriorate as sharply as the performance of the boosting approach in these noisier settings. The MSB of $\hat{\varphi}^{\text{BC}}(x)$ is lower than or equal to the MSB of the boosting estimator in all but the Friedman dataset. Additionally, the MSPE of $\hat{\varphi}^{\text{BC}}(x)$ is lower than the MSPE of the boosting estimator in every dataset, including the Friedman dataset.

\begin{table}[h]
    \centering
    \begin{tabular}{ccccccc}
         \toprule
         & \multicolumn{3}{c}{MSB} & \multicolumn{3}{c}{MSPE} \\
         \cmidrule(r){2-4} \cmidrule(l){5-7}
         Dataset (Noised)         & RF     & Boost & $\hat{\varphi}^{\text{BC}}(x)$ & RF & Boost & $\hat{\varphi}^{\text{BC}}(x)$\\
         \cline{1-7}
         Baseline       & 0.008 & 0.019 & 0.009 & 106.536 & 117.180 & 108.761\\
         Linear           & 0.018 & 0.019 & 0.016 & 106.505 & 117.059 & 108.686\\
         Step             & 2.166 & 1.295 & 0.985 & 109.035 & 118.516 & 110.748\\
         Exponential & 1.291 & 1.617 & 0.522 & 165.438 & 191.892 & 168.375\\
         Friedman     & 5.142 & 2.081 & 3.441 & 112.206 & 119.532 & 113.898\\
         \bottomrule
    \end{tabular}
    \caption{Mean squared bias and mean squared prediction error of the standard random forest, the bias-corrected random forest based on boosting, and our bias-corrected random forest $\hat{\varphi}^{\text{BC}}(x)$ for each noised synthetic dataset.}
    \label{tab:additional_bias_results}
\end{table}

\subsection{Conditional Prediction Interval Estimation}
\label{sec:interval_sim}
Next, we compare our prediction interval estimator $\widehat{\text{PI}}_{\alpha}(x)$ to the estimators obtained by quantile regression forests \citep{meinshausen2006}, generalized random forests \citep{Athey2019}, conformal inference \citep{Lei2014, Lei2018, Johansson2014}, and the unweighted out-of-bag approach of \citet{zhang2019}. For the conformal inference estimator, we specifically used the locally weighted split conformal inference procedure proposed by \citet{Lei2018}, which is a special case of the split conformal inference approach that attempts to account for residual heterogeneity across the covariate space by using standardized residuals for the conformity scores. Via simulation, we evaluate these methods with respect to three metrics: coverage rate, interval width, and qualitative behavior. In each simulation, we randomly sampled 1,000 training units and 1,000 test units, and applied each method to construct 95\% prediction intervals for the test units. We repeated this procedure 1,000 times for each of the following datasets.

\begin{description}
\item[Linear:] $X \simiid \text{Unif}[-1, 1]^{50}$, and $Y \simind \mathcal{N}\left(X_1, 4\right)$.
\item[Clustered:] $X \in [0, 1]^{10}$ is drawn i.i.d. from a population consisting of five distinct, roughly equally sized clusters, with no overlap between clusters. $Y$ is independently drawn from a normal distribution with mean and variance determined by the cluster to which $X$ belongs. The response means and variances within the clusters are $\{(0, 1), (40, 4), (80, 9), (120, 16), (160, 25)\}$. See \citet{maitra2010} for details. The data were generated using the {\ttfamily MixSim} package \citep{MixSimR}.
\item[Step:] With probability 0.05, $X \simiid \text{Unif}\left([-1, 0] \times [-1, 1]^{9}\right)$; else, $X \simiid \text{Unif}\left([0, 1] \times [-1, 1]^{9}\right)$. $Y \simind \mathcal{N}\left(20\cdot\mathbbm{1}(X_1 > 0), 4\right)$.
\item[Friedman:] $X \simiid \text{Unif}[-1, 1]^{10}$, and $Y \simind \mathcal{N}\left(10\sin(\pi X_1X_2) + 20\left(X_3 - 1/2\right)^2 + 10X_4 + 5X_5, 1\right)$.
\item[Parabola:] With probability 0.05, $X \simiid \text{Unif}\left([-1, -1/3] \times [-1, 1]^{39}\right)$; with probability 0.9, $X \simiid \text{Unif}\left([-1/3, 1/3] \times [-1, 1]^{39}\right)$; and with probability 0.05, $X \simiid \text{Unif}\left([1/3, 1] \times [-1, 1]^{39}\right)$. $Y \simind \mathcal{N}\left(0, X_1^4\right)$.
\item[2D:] $X \simiid \text{Unif}[-1, 1]^{50}$, and $Y \simind \mathcal{N}\left(5X_1, 4(X_2 + 2) ^ 2\right)$.
\end{description}
In addition to conducting simulations on the above synthetic datasets, we also randomly partitioned each of the Boston, Abalone, and Servo benchmark datasets into training and test sets using the same train-test ratios as in Section \ref{sec:bias_sim} and estimated prediction intervals for the test points. We repeated this 1,000 times for each of the three benchmark datasets.

Table \ref{tab:interval_results} shows the average coverage rate of each method in each simulation, with average interval widths shown in parentheses. Overall, all five methods performed fairly well with respect to these two metrics. However, it is notable that generalized random forest intervals were the widest and tended to heavily over-cover in nearly every dataset; they were particularly wide in the Clustered dataset. While our $\widehat{\text{PI}}_{\alpha}(x)$ intervals also over-covered in the Freidman and Parabola datasets, they were mostly no wider than other methods' intervals that covered at the desired 95\% rate. For example, our method produced narrower intervals than conformal inference and the unweighted out-of-bag approach in the Friedman dataset despite having a higher coverage rate. Additionally, our method produced narrower intervals than the unweighted out-of-bag method despite having a higher coverage rate in almost half the datasets. Finally, quantile regression forests noticeably under-covered in the Clustered dataset even though the intervals were wider than $\widehat{\text{PI}}_{\alpha}(x)$ on average.

\begin{table}
    \centering
    \begin{tabular}{cccccc}
         \toprule
         Dataset     &      QRF           &        GRF         &         Split         &         OOB          & $\widehat{\text{PI}}_{\alpha}(x)$\\
         \cline{1-6}
	Linear        & 0.949 (8.04)    &  0.952 (8.11)    &  0.950 (8.32)    &  0.949 (7.92)     &  0.948 (7.95)\\
	Clustered  & 0.930 (15.15)  &  0.966 (41.27)  &  0.950 (15.72)    &  0.949 (15.83)   &  0.945 (13.94)\\
	Step          & 0.944 (9.15)    &  0.962 (12.10)  &  0.951 (9.17)    &  0.949 (8.28)     &  0.945 (8.17)\\
	Friedman  & 0.992 (36.27)  &  0.991 (45.50)  &  0.950 (22.19)   &  0.949 (23.34)  &  0.969 (22.01)\\
       	Parabola   & 0.960 (0.82)    &  0.960 (0.84)    &  0.951 (0.79)     &  0.949 (0.84)    &  0.967 (0.83)\\
	2D             & 0.957 (18.07)  &  0.962 (18.87)  &   0.951 (16.99) &  0.948 (17.22)   &  0.951 (17.25)\\
         Boston      & 0.981 (15.57)  &  0.994 (23.92)  &  0.951 (13.21)  &  0.946 (12.64)   &  0.947 (11.16)\\
         Abalone    & 0.969 (7.95)    &  0.982 (9.21)    &  0.950 (8.39)    &  0.950 (9.11)     &  0.949 (8.17)\\
         Servo       & 0.951 (24.42)   &  0.985 (37.29)  &  0.961 (27.33)  &  0.943 (21.13)  &  0.946 (18.85)\\
         \bottomrule
    \end{tabular}
    \caption{Average coverage rates and widths of 95\% prediction intervals constructed by quantile regression forests, generalized random forests, split conformal inference, the unweighted out-of-bag method, and $\widehat{\text{PI}}_{\alpha}(x)$.}
    \label{tab:interval_results}
\end{table}

\begin{figure}
    \centering
    \includegraphics[width = 0.496\linewidth]{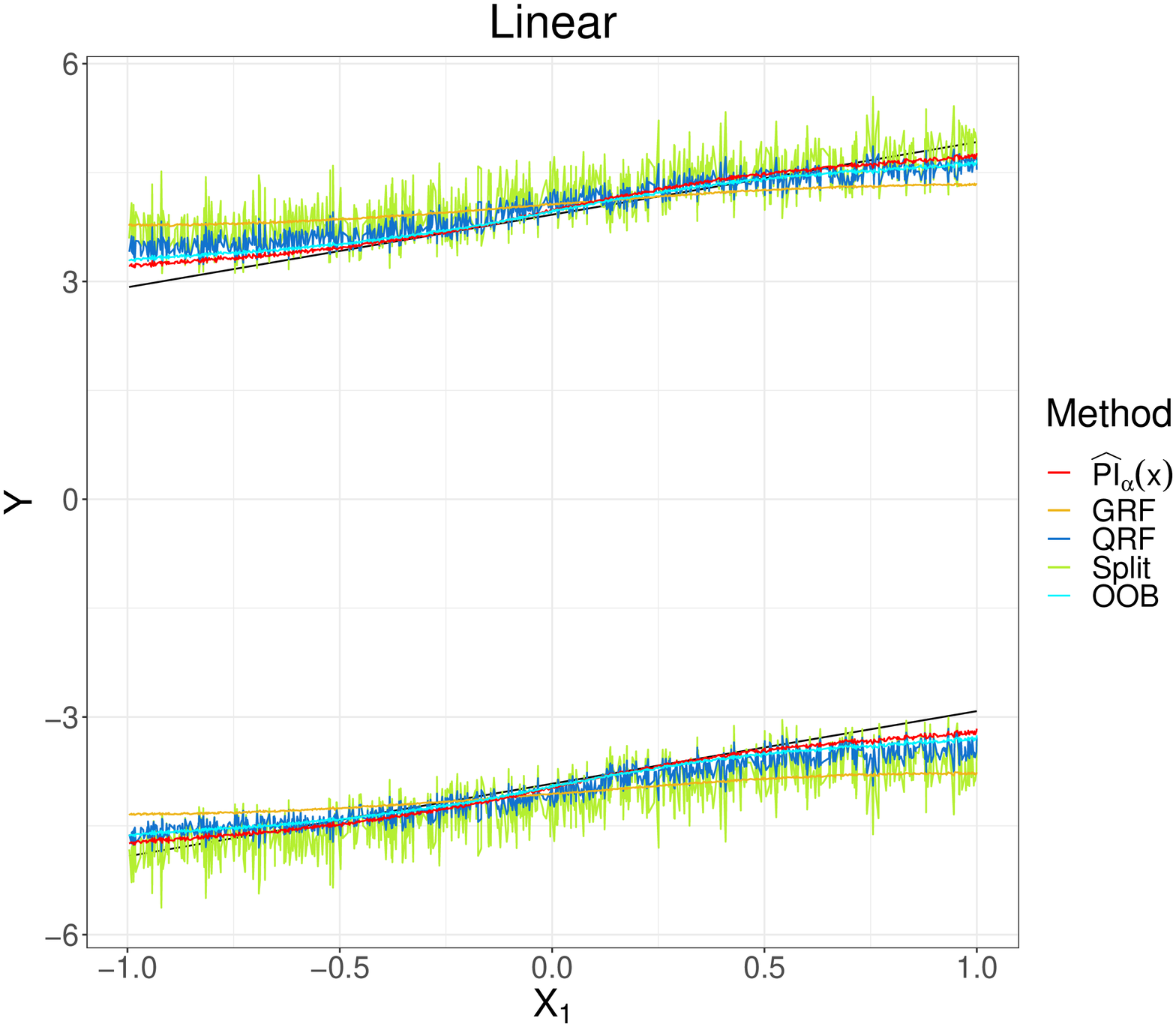}
    \includegraphics[width = 0.496\linewidth]{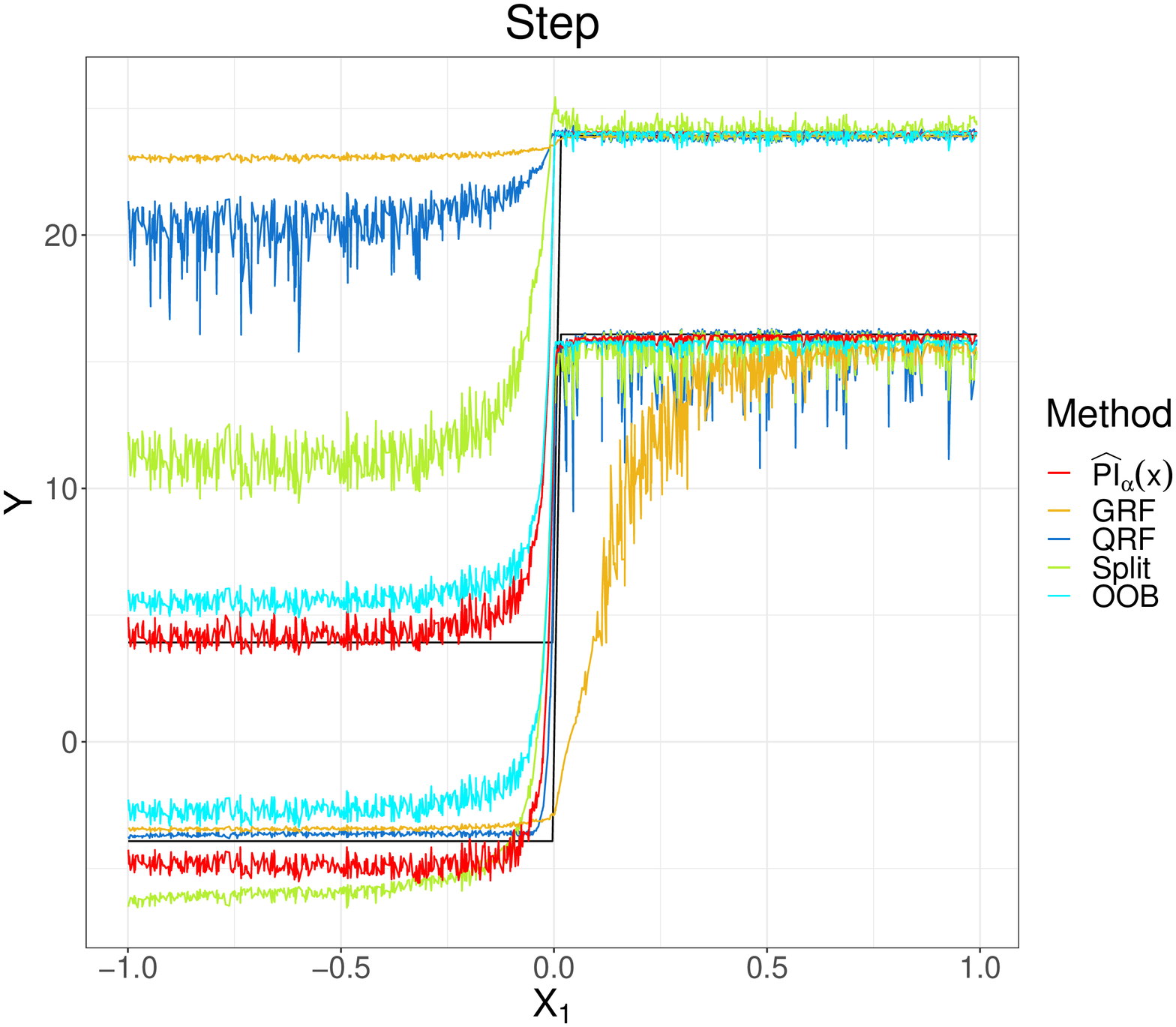}
    \includegraphics[width = 0.496\linewidth]{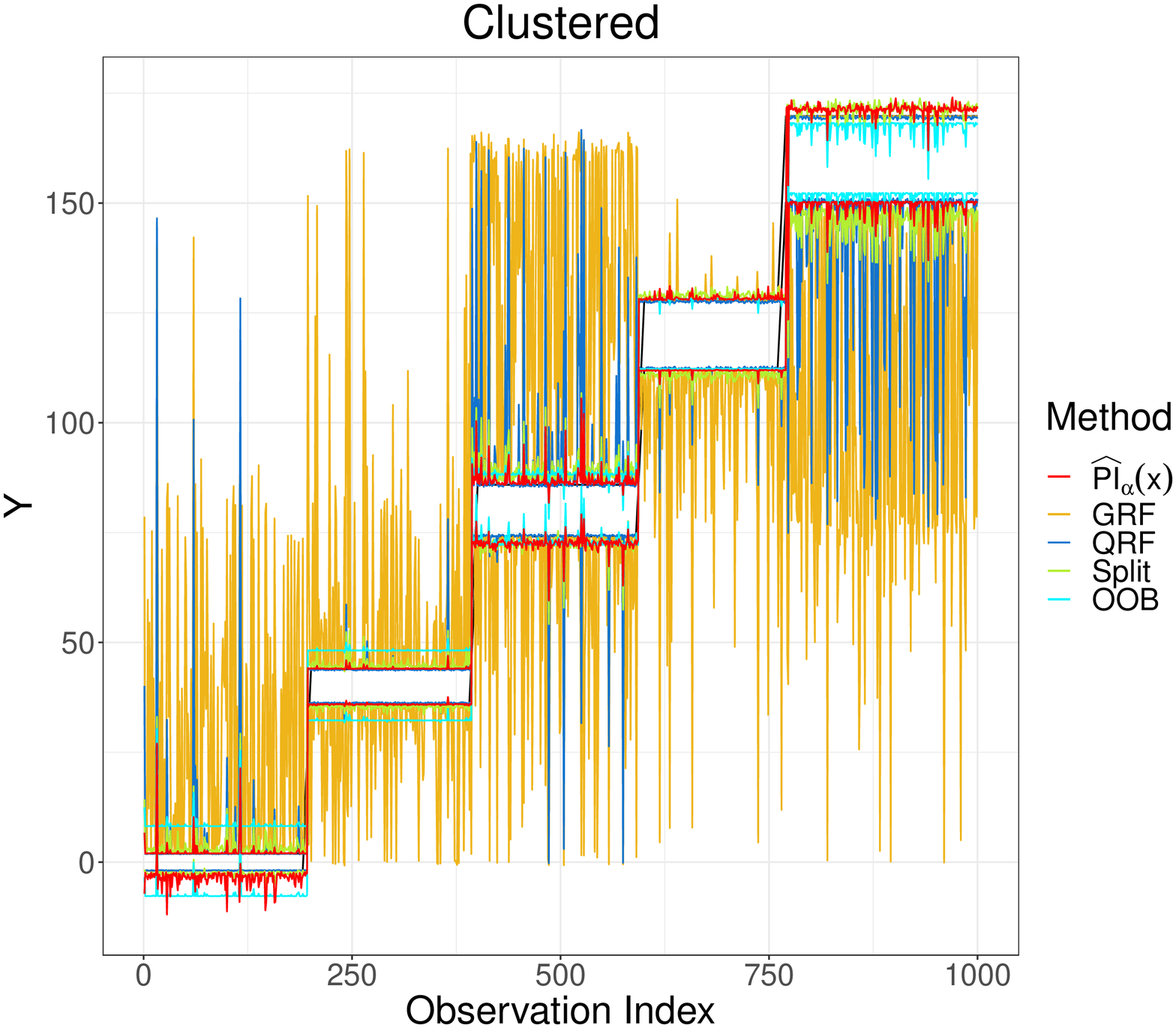}
    \includegraphics[width = 0.496\linewidth]{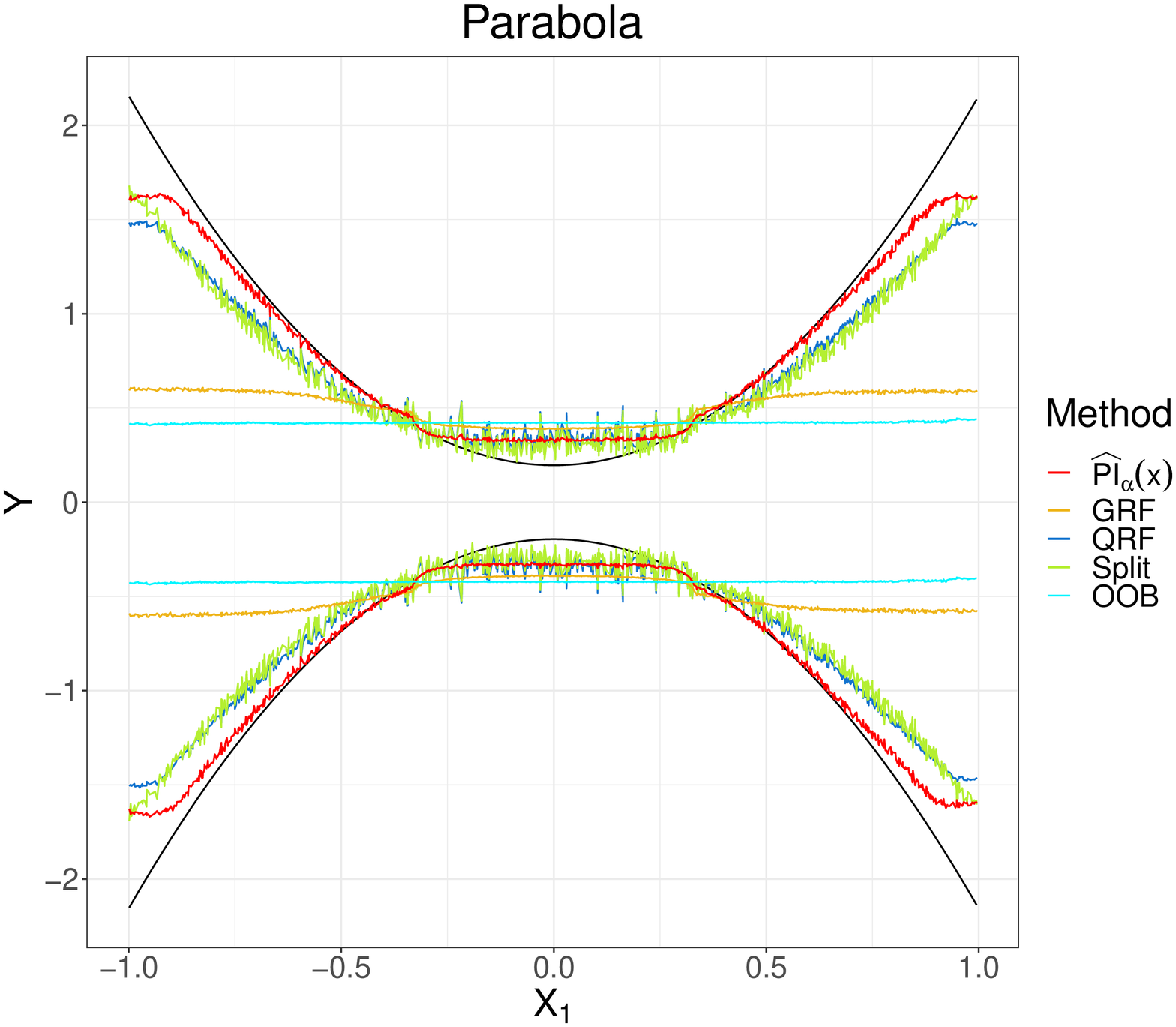}
    \caption{Average upper and lower bounds of 95\% prediction intervals constructed by our method $\widehat{\text{PI}}_{\alpha}(x)$, generalized random forests, quantile regression forests, split conformal inference, and the unweighted out-of-bag method for the Linear, Step, Parabola, and Clustered datasets (clockwise from top left) over 1,000 simulation repetitions. The true target conditional response quantiles are shown in black.}
    \label{fig:interval_behavior}
\end{figure}

Table \ref{tab:interval_results}, however, reports only \textit{unconditional} coverage rates and interval widths, computed over the entire test sample. Although these unconditional metrics are important, researchers in practice often seek prediction intervals with good coverage rates and widths \textit{conditionally}---that is, given a specific test observation of interest. To better evaluate how each method performs \textit{conditionally}, we plot in Figure \ref{fig:interval_behavior} the average estimated conditional response quantiles against the true conditional quantiles for the Linear, Clustered, Step, and Parabola datasets. Overall, $\widehat{\text{PI}}_{\alpha}(x)$ captured the nuances in the structure of the data better than the other estimators. In all four datasets, $\widehat{\text{PI}}_{\alpha}(x)$ best tracked the changes in the conditional quantiles across the covariate space. Only $\widehat{\text{PI}}_{\alpha}(x)$ correctly estimated the upper quantile when $X_1 < 0$ in the Step dataset. Moreover, generalized random forests did not capture the strong curvature of the quantiles in the Parabola dataset, quantile regression forests and generalized random forests produced erratic intervals in the Clustered dataset, and the out-of-bag approach of \citet{zhang2019} failed to reflect any heterogeneity in the Parabola and Clustered datasets. Additionally, while all methods performed fairly well in the Linear dataset, ours exhibited the least bias at the boundaries of the covariate space.

We believe that at least some of the undesirable behaviors exhibited by generalized random forests and quantile regression forests in these simulations can be attributed to the methods' use of response quantiles instead of out-of-bag prediction error quantiles. For example, the quantile regression variant of generalized random forests partitions the covariate space based on the empirical quantiles of the training responses, so it is less able to detect changes in the conditional response quantiles in low-density regions of the covariate space. This can be seen in the Step and Parabola datasets. Perhaps more importantly, after their trees are grown, both generalized random forests and quantile regression forests use training responses directly to impute the conditional response distribution, so their prediction intervals are more sensitive to sharp discontinuities in the conditional response distribution that are not accurately delineated by their tree-growing algorithms. This can be seen in the Step and Clustered datasets.

Unlike generalized random forests and quantile regression forests, conformal inference and the out-of-bag approach of \citet{zhang2019} do use prediction error quantiles to construct prediction intervals. But neither method weights these quantiles based on the training observations' similarity to the test observation. Conformal inference directly uses the empirical distribution of conformity scores, weighting each score equally. Similarly, the out-of-bag approach of \citet{zhang2019} uses the unweighted quantiles of the out-of-bag errors instead of weighting them based on cohabitation frequency or some other similarity metric. Thus, by construction, the intervals of \citet{zhang2019} have the same width for all test observations and ignore any heterogeneity in the shape of the conditional response distribution across the covariate space, as seen in the Clustered and Parabola datasets.

\begin{figure}
    \centering
    \includegraphics[width = 0.496\linewidth]{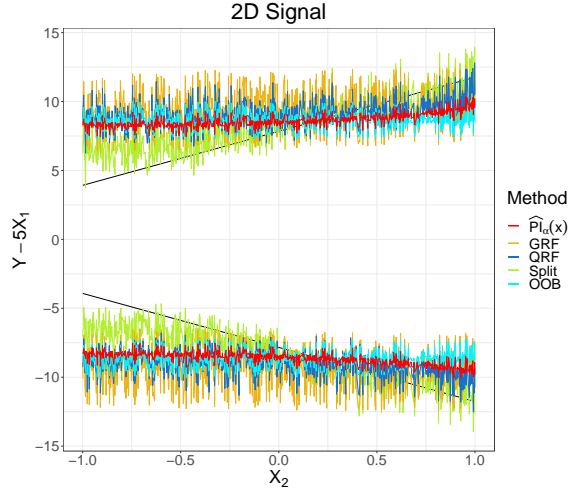}
    \caption{Average upper and lower bounds of 95\% prediction intervals constructed by our method $\widehat{\text{PI}}_{\alpha}(x)$, generalized random forests, quantile regression forests, split conformal inference, and the unweighted out-of-bag method for the 2D dataset over 1,000 simulation repetitions. The true target conditional response quantiles are shown in black.}
    \label{fig:2d}
\end{figure}

Our method of prediction interval estimation avoids both of these pitfalls. Rather than using response quantiles, as quantile regression forests and generalized random forests do, our method uses out-of-bag prediction error quantiles, thus more fully leveraging the predictive power of the random forest. Additionally, our method weights the errors by how closely the training units resemble the test point of interest, unlike conformal inference and the out-of-bag approach of \citet{zhang2019}. Of course, this is not to say that our method is uniformly best. For example, although they may not estimate conditional response quantiles as well or cover at the desired rate conditionally in some settings, conformal inference prediction intervals are guaranteed to cover at the desired rate unconditionally even in finite samples. Additionally, our method qualitatively performs as poorly as the others in the 2D dataset, where the conditional response mean and variance depend on separate covariates (Figure \ref{fig:2d}). This structure is challenging for many tree-based estimators, which usually split based on heterogeneity in only one aspect of the conditional response distribution.

\begin{figure}
    \centering
    \includegraphics[width = 0.496\linewidth]{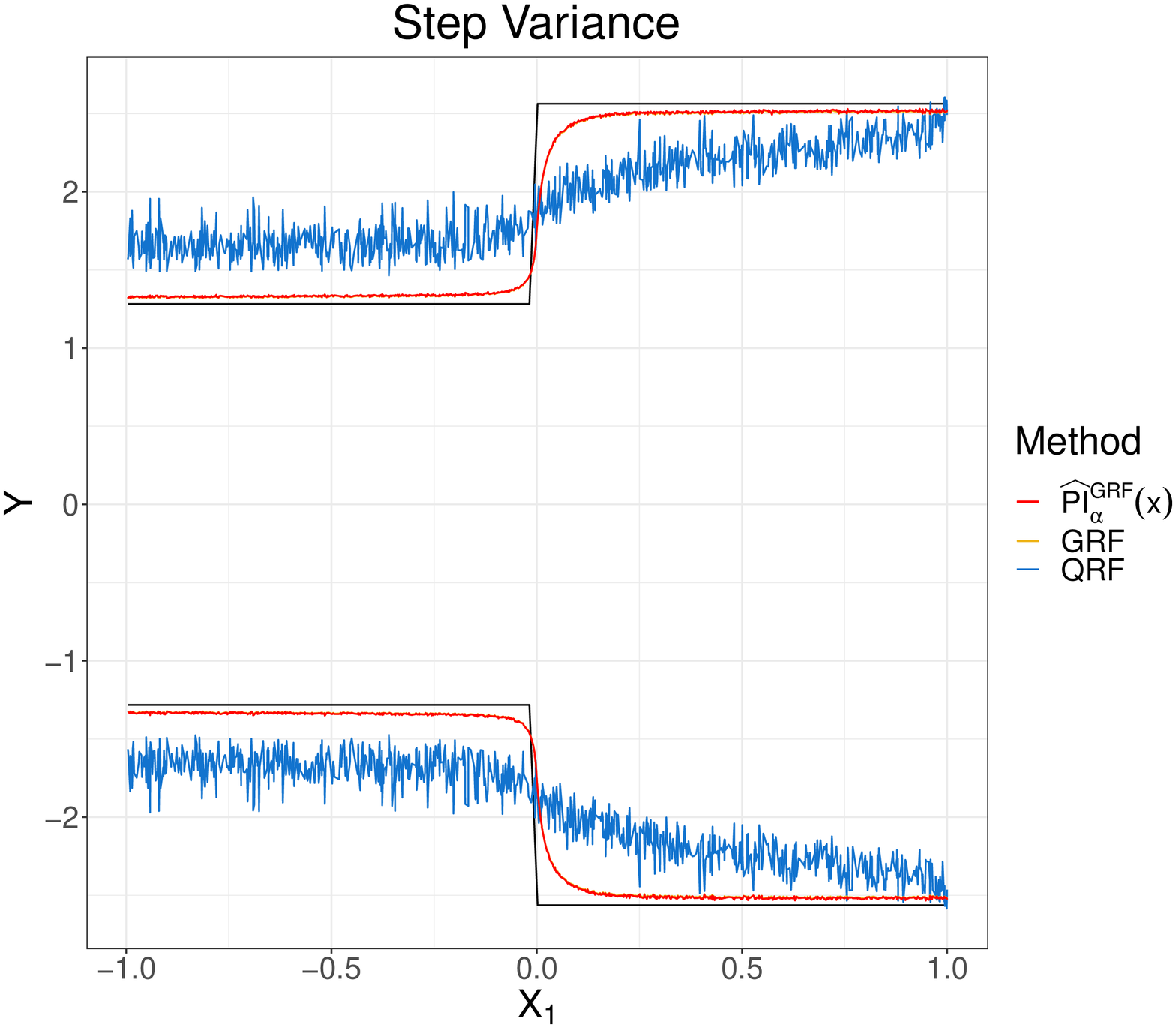}
    \includegraphics[width = 0.496\linewidth]{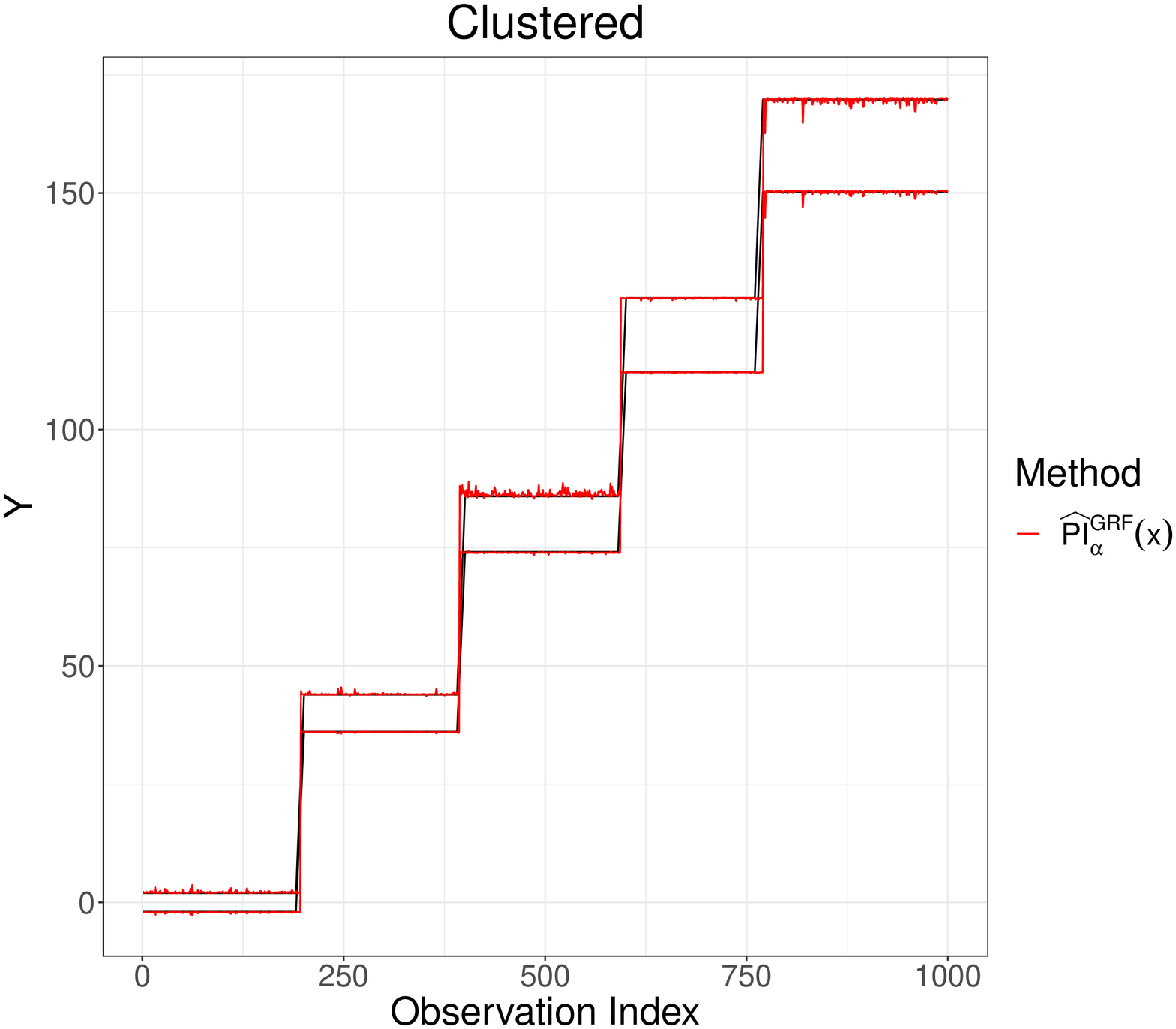}
    \caption{Average upper and lower bounds of prediction intervals constructed by our method $\widehat{\text{PI}}_{\alpha}^{\text{GRF}}(x)$ applied to generalized random forests for the simulation of \citet{Athey2019} (left) and the Clustered dataset (right) over 1,000 simulation repetitions. Average upper and lower bounds of other methods' prediction intervals are included for comparison. The true target conditional response quantiles are shown in black.}
    \label{fig:grf}
\end{figure}

To illustrate the generality of our framework, we also applied our method to the generalized random forest tree construction algorithm. \citet{Athey2019} show via simulation that generalized random forests outperform quantile regression forests when the response variance follows a step function but the mean response is constant. This is because quantile regression forests grow trees using the CART algorithm, which is sensitive only to mean shifts. Our method as applied in the simulations thus far has also employed the CART algorithm and therefore has also inherited this limitation. But our method is well-defined regardless of the underlying tree-growing algorithm. So we can easily apply our method to the generalized random forest algorithm, with minor deviations detailed in Appendix \ref{ap:grf}, instead; we denote this adaptation by $\widehat{\text{PI}}_{\alpha}^{\text{GRF}}(x)$. Figure \ref{fig:grf} replicates the simulation of \citet{Athey2019} comparing quantile regression forests and generalized random forests. $\widehat{\text{PI}}_{\alpha}^{\text{GRF}}(x)$, added in red, performs identically to generalized random forests in this setting. More generally, we find that $\widehat{\text{PI}}_{\alpha}^{\text{GRF}}(x)$ estimates the conditional response quantiles as well as or better than generalized random forests in all datasets used earlier. Full simulation results are in Appendix \ref{ap:grf}, but, as a notable example, a comparison between Figure \ref{fig:grf} and Figure \ref{fig:interval_behavior} shows that $\widehat{\text{PI}}_{\alpha}^{\text{GRF}}(x)$ outperforms every other method in the Clustered dataset.

\section{Theoretical Result}
\label{sec:theory}
In this section, we propose a similar but more stringent estimator of $F_E(e \mid x)$ and prove that it is uniformly consistent. In addition to regularity assumptions on the tree construction procedure that have become somewhat standard in recent literature, the most notable difference between $\hat{F}_{E}(e \mid x)$ as defined in (\ref{eq:mainresult}) and the estimator discussed here is that the latter uses two random forests fit on disjoint subsets of the training data. One random forest produces weights, and the other produces out-of-sample prediction errors. We view this broadly as a strengthening of the independence conditions that motivated the out-of-bag construction of $\hat{F}_E(e \mid x)$.

While we do not prove that our practical estimator $\hat{F}_E(e \mid x)$ is uniformly consistent, we nonetheless consider it noteworthy for two reasons. From an applied perspective, it is more data-efficient than our stringent estimator and appears to perform just as well in general; this can be seen in simulations implementing our stringent estimator in Appendix \ref{ap:stringent}. From a theoretical perspective, the contrast between $\hat{F}_E\left(e \mid x\right)$ and the estimator introduced in this section highlights aspects of the dependence structure of random forests that we believe merit further investigation. Future research into such topics may close the gap between the two versions of our estimator and contribute more generally to a deeper understanding of tree-based algorithms.

\subsection{Stringent Estimator of the Conditional Prediction Error Distribution}
\label{sec:alg}
The algorithm for computing our more stringent estimator of $F_E\left(e \mid x\right)$ is outlined below. Note that, in what follows, we redefine some earlier notation rather than introduce new symbols to reduce notational complexity. An effort has been made to be explicit whenever notation is redefined.
\begin{enumerate}
    \item Partition the training data evenly into three subsets, arbitrarily labeled $\mathcal{I}$, $\mathcal{J}$, and $\mathcal{K}$. Let $n$ denote the sample size of each subset, as opposed to the sample size of the full training set.
    \item Grow one random forest with $B$ trees using $\mathcal{I}$. Label it the ``first random forest.''
    \item Grow one random forest with $B$ trees using $\mathcal{J}$ and the covariates in $\mathcal{K}$. In other words, do not consider the responses of the units in $\mathcal{K}$ when splitting tree nodes. Label this random forest the ``second random forest.''
    \item Compute the errors of the first random forest's predictions of the $n$ units in $\mathcal{K}$:
    \[
    E_i := Y_i - \hat{\varphi}(X_i),
    \]
    where $\hat{\varphi}$ denotes the first random forest estimator.
    \item For a target $x$, compute the weight of each of the $n$ units in $\mathcal{K}$ given by the second random forest:
    \begin{equation}
    \label{eq:honestvi}
    v_i(x) := \frac{1}{B}\sum_{b = 1}^{B}\frac{\#\{Z_i \in \mathcal{D}^*_{n,b}\}\mathbbm{1}(X_i \in R_{\ell(x, \theta_b)})}{\sum_{j = 1}^{n}\#\{Z_j \in \mathcal{D}^*_{n,b}\}\mathbbm{1}(X_j \in R_{\ell(x, \theta_b)})},
    \end{equation}
    where $\mathcal{D}^*_{n,b}$ denotes the bootstrap set of units from $\mathcal{K}$ whose covariates were used in the construction of the $b^{\text{th}}$ tree of the second random forest, and $R_{\ell(x, \theta_b)}$ denotes the rectangular subspace corresponding to the terminal node of the $b^{\text{th}}$ tree of the second random forest in which $x$ falls. Recall that $v_i(x)$ is a random variable based in part on $X_i$, hence the subscript $i$.
    \item Letting $\hat{F}_E(e \mid x)$ now denote our stringent estimator rather than the estimator given by (\ref{eq:mainresult}) in Section \ref{sec:prederror}, define
    \begin{equation}
    \label{eq:stringent}
    \hat{F}_E(e \mid x) := \sum_{i = 1}^{n}v_i(x)\mathbbm{1}(E_i \leq e).
    \end{equation}
\end{enumerate}
Step 3 of our procedure is similar to the honest double-sample regression tree algorithm of \citet{Wager2018}, but here the training data are split into subsets before resampling or subsampling. One way to grow the second random forest is to use the covariates in $\mathcal{J}$ and $\mathcal{K}$ to determine the set of eligible splits, then choose the eligible split that optimizes some empirical objective of the responses in $\mathcal{J}$ only. Another approach is to grow trees using only data from $\mathcal{J}$, then prune terminal nodes that do not contain any units from $\mathcal{K}$.

\subsection{Consistency}

Because the number of trees can be made arbitrarily large given enough computational power, we take the approach of \citet{Scornet2015} and prove that the limiting version (as $B \to \infty$) of $\hat{F}_E(e \mid x)$ given by (\ref{eq:stringent}) is consistent as $n \to \infty$. This is justified by the law of large numbers. We do so under the following set of assumptions, many of which are from \citet{meinshausen2006}. First, we make an assumption about the covariate distribution.
\begin{assumption}
\label{as:uniform}
$X$ has the uniform distribution over $[0, 1]^p$.
\end{assumption}
\noindent Assumption \ref{as:uniform} is largely for notational convenience. More generally, one could assume that the density of $X$ is positive and bounded.

We also make a set of assumptions about the way the observations in $\mathcal{K}$ are used in the construction of the second random forest. For any generic tree in the second random forest grown with parameter vector $\theta$, let $k_{\theta}(\ell) := \left|\{Z_i \in \mathcal{D}^*_{n}:X_i \in R_{\ell(x, \theta)}\}\right|$ denote the number of units from its bootstrap sample $\mathcal{D}^*_n$ of $\mathcal{K}$ in its terminal node containing $x$.
\begin{assumption}
\label{as:propobs}
\leavevmode
\begin{enumerate}[label=(\alph*)]
\item The proportion of observations from $\mathcal{D}^*_n$ in any given node, relative to all observations from $\mathcal{D}^*_n$, is decreasing in $n$---that is, $\max_{\ell, \theta}k_{\theta}(\ell) = o(n)$. The minimum number of observations from $\mathcal{D}^*_n$ in a node is increasing in $n$---that is, $1 / \min_{\ell, \theta}k_{\theta}(\ell) = o(1)$.
\item The probability that variable $m \in \{1, \ldots, p\}$ is chosen for a given split point is bounded from below for every node by a positive constant.
\item When a node is split, the proportion of observations belonging to $\mathcal{D}^*_n$ in the original node that fall into each of the resulting sub-nodes is bounded from below by a positive constant.
\end{enumerate}
\end{assumption}
\noindent The conditions in Assumption \ref{as:propobs} are adapted from assumptions used to prove consistency of quantile regression forests \citep{meinshausen2006}. Tree construction algorithms that satisfy these properties or variants of them have been referred to in recent random forest literature as ``regular,'' ``balanced,'' or ``random-split'' \citep{Wager2018, Athey2019, Friedberg2019}.

Next, we assume that the distribution of prediction errors is sufficiently smooth.
\begin{assumption}
\label{as:error}
$F_E(e \mid X = x)$ is Lipschitz continuous with parameter $L$. That is, for all $x, x' \in [0, 1]^p$,
\[
\sup_{e \in \mathbb{R}}\left|F_E(e \mid X = x) - F_E(e \mid X = x')\right| \leq L\|x - x'\|_1.
\]
\end{assumption}
\noindent As \citet{Wager2018} note, all existing results on pointwise consistency of random forests have required an analogous smoothness condition in the distribution of interest, including \citet{biau2012}, \citet{meinshausen2006}, and \citet{Wager2018}.

Additionally, we assume that the distribution of prediction errors is strictly monotone so that consistency of quantile estimates follows from consistency of distribution estimates.
\begin{assumption}
\label{as:monotone}
$F_E(e \mid X = x)$ is strictly monotone in $e$ for all $x \in [0, 1]^p$.
\end{assumption}

We also assume that the random forest is stable in the following sense.
\begin{assumption}
\label{as:consistency}
There exists a function $\varphi(\cdot)$ such that $\hat{\varphi}(X) - \varphi(X) \toProb 0$ as $n \to \infty$, with $-\infty < \varphi(X) < \infty\ a.s.$
\end{assumption}
\noindent It may help one's intuition to imagine that $\varphi(X) = \mathbb{E}[Y \mid X]$, in which case Assumption \ref{as:consistency} simply states that the random forest is consistent. But $\varphi(X)$ need not be the conditional mean response. Note also that Assumption \ref{as:consistency} does not require stability as defined by \citet{buhlmann2002}, since here the convergence does not have to be pointwise. Stability---and, in particular, consistency---of random forests is an ongoing area of research. \citet{Scornet2015} prove consistency of the original random forest algorithm of \citet{breiman2001} when the underlying data follow an additive regression model. \citet{wager2015adapt} prove consistency of adaptively grown random forests, including forests built using CART-like algorithms, in high-dimensional settings.

Finally, we make an assumption about the behavior of the weights given by the second random forest relative to the predictions of the first random forest. For any $\delta > 0$, define the event $\MM := \{\left|\hat{\varphi}(X_i) - \varphi(X_i)\right| < \delta\}$. We say that $\delta$-stability of the $i^{\text{th}}$ unit has been realized if and only if $\MM$ holds.
\begin{assumption}
\label{as:weight}
For all $x \in [0, 1]^p$, there exists $\delta_0 > 0$ such that, for any $\delta \in \left(0, \delta_0\right)$, $\EE[\vx{i} \mid \MM] = O(n^{-1})$ and $\EE[\vx{i} \mid \neg\MM] = O(n^{-1})$.
\end{assumption}
\noindent Assumption \ref{as:weight} further characterizes the stability of the random forest and the underlying population distribution. It states that the expected out-of-bag weight of the $i^{\text{th}}$ observation in $\mathcal{K}$---which, recall from its definition in (\ref{eq:honestvi}), is a random variable in $X_i$ and other quantities---is of order $1 / n$ whether $\delta$-stability has been realized for the observation or not. The expected values are taken over all training units and all random parameters governing the sample-splitting and tree-growing mechanisms. Notice that Assumption \ref{as:weight} is satisfied if $\EE\left[\vx{i} \mid \MM\right] > \EE\left[\vx{i} \mid \neg\MM\right]$ and Assumption \ref{as:consistency} holds since the weights must be nonnegative and $\EE[\vx{i}] = 1 / n$. Note also that the bounding constant can vary by $\delta \in (0, \delta_0)$.

Under these assumptions, we prove in Appendix \ref{ap:proof} that $\hat{F}_E(e \mid x)$ is a uniformly consistent estimator for the true conditional prediction error distribution $F_E(e \mid x)$.

\begin{theorem2}
\label{th:main}
Under Assumptions \ref{as:uniform}-\ref{as:weight},
\[
\sup_{e \in \mathbb{R}}\left|\hat{F}_E(e \mid x) - F_E(e \mid x)\right| \toProb 0,\hspace{20pt}n \to \infty
\]
pointwise for every $x \in [0, 1]^p$.
\end{theorem2}

\section{Conclusion}
\label{sec:conclusion}

We propose a unified framework for random forest prediction error estimation based on a novel estimator for the conditional prediction error distribution. Under this framework, useful uncertainty metrics can be estimated by simply plugging in the estimated conditional prediction error distribution. By contrast, these quantities previously each had to be estimated by different, and in some cases not obviously compatible, algorithms. We demonstrate the unified nature of our approach by deriving, to our knowledge, the first estimator for the conditional mean squared prediction error of random forests, as well as estimators for conditional bias and conditional prediction intervals that are competitive with, and in some cases outperform, existing methods.

We believe that one advantage of our framework is its generality. While this paper discusses our work primarily in the context of CART, our estimators can be readily adapted to other bagged, tree-based estimators with different splitting criteria and subsampling rules, as demonstrated by the adaptation of our method to generalized random forests in Section \ref{sec:interval_sim}. The weighting scheme we propose can also be naturally tailored to specific needs. For example, the weights can be modified to count cohabitation in non-terminal nodes if more stability is needed. More broadly, we believe that our general approach of weighting out-of-sample prediction errors by their similarity to the test point of interest with respect to the estimator is applicable to a wide range of estimators with suitably defined metrics for similarity, even those not based on decision trees. While beyond the scope of this paper, future work into such extensions may prove fruitful.

\acks{The authors gratefully acknowledge the Pomona College Summer Undergraduate Research Program and the Pomona College Kenneth Cooke Summer Research Fund for their support of this research. The authors thank the action editor and three anonymous reviewers for their helpful feedback. This material is based upon work supported by the National Science Foundation under Grant No. 1745640.}

\newpage

\appendix

\section{Simulation Details and Additional Results}
\label{ap:sim}

\subsection{Parameter Settings of Main Results}
\label{ap:params}

We ran all bias and prediction interval simulations except our replication of the \citet{Athey2019} prediction interval simulation (Figure \ref{fig:grf}, left panel) with the following parameters. Each forest consisted of 1,000 trees. The minimum node size parameter for all forests was set to 5. We set the number of covariates randomly sampled as candidates at each split to $\max\{\left \lfloor{p / 3}\right \rfloor, 1\}$, where $p$ is the number of covariates. We used the default sample-splitting regime for generalized random forests given in the {\ttfamily grf} package in {\ttfamily R}: Half of the training data were used to build each tree, with half of those units held out for honest tree growth. For our replication of the \citet{Athey2019} simulation, we used the same parameter settings as above except we set the number of covariates randomly sampled as candidates at each split to $\min\{\left \lceil{\sqrt{p}} + 20\right \rceil, p\}$, following \citet{Athey2019}.

\subsection{Stringent Estimator Implementation Details and Results}
\label{ap:stringent}

We implemented and evaluated the performance of the stringent versions of our bias and prediction interval estimators as described in Section \ref{sec:alg}, with two slight modifications. In the third step of our procedure, we did not use any data from $\mathcal{K}$ to construct the second random forest. Because of this, we could not guarantee that each terminal node of the second random forest contained a unit from $\mathcal{K}$, so we computed $v_i(x)$ in the fifth step of our procedure by counting the number of times the $i^{\text{th}}$ unit in $\mathcal{K}$ was a cohabitant of $x$ and dividing by the total number of times any unit in $\mathcal{K}$ was a cohabitant of $x$:
\[
v_i(x) = \frac{\sum_{b = 1}^{B}\mathbbm{1}\left(X_i \in R_{\ell\left(x, \theta_b\right)}\right)}{\sum_{j = 1}^{n}\sum_{b = 1}^{B}\mathbbm{1}\left(X_j \in R_{\ell\left(x, \theta_b\right)}\right)}.
\]
We believe that these deviations are minor and that this implementation reflects the major features that differentiate the stringent version of our estimator from the practical version, particularly the independence relations enforced by growing two random forests on disjoint subsets of data. Because our stringent estimator splits the training set into three subsets, which we expected would reduce efficiency, we evaluated our stringent estimator on the synthetic datasets using both the original training sample sizes (200 for the bias simulations and 1,000 for the prediction interval simulations) and triple the original training sample sizes; we were, of course, unable to similarly augment the benchmark datasets.

\begin{figure}
    \centering
    \includegraphics[width = 0.496\linewidth]{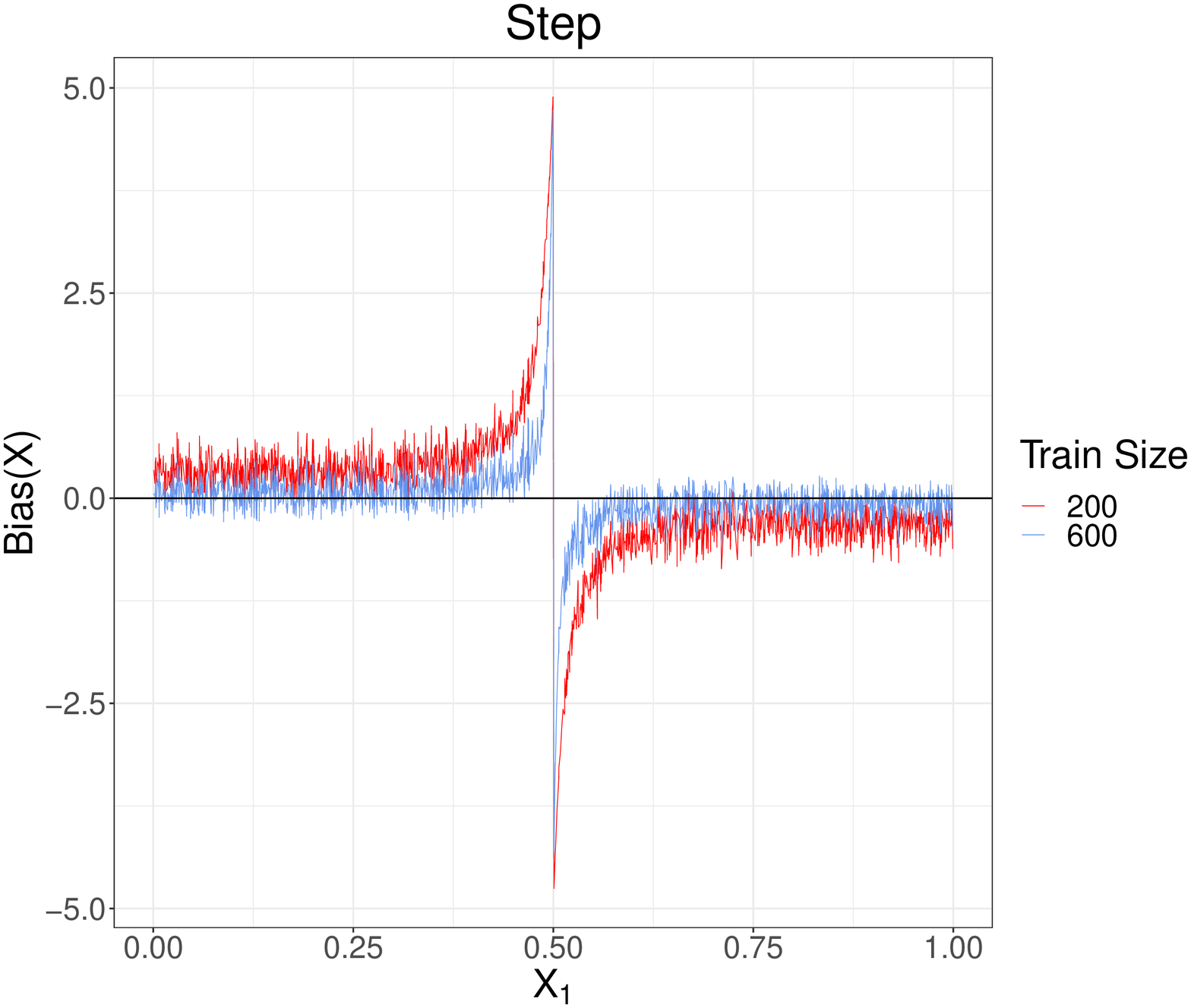}
    \includegraphics[width = 0.496\linewidth]{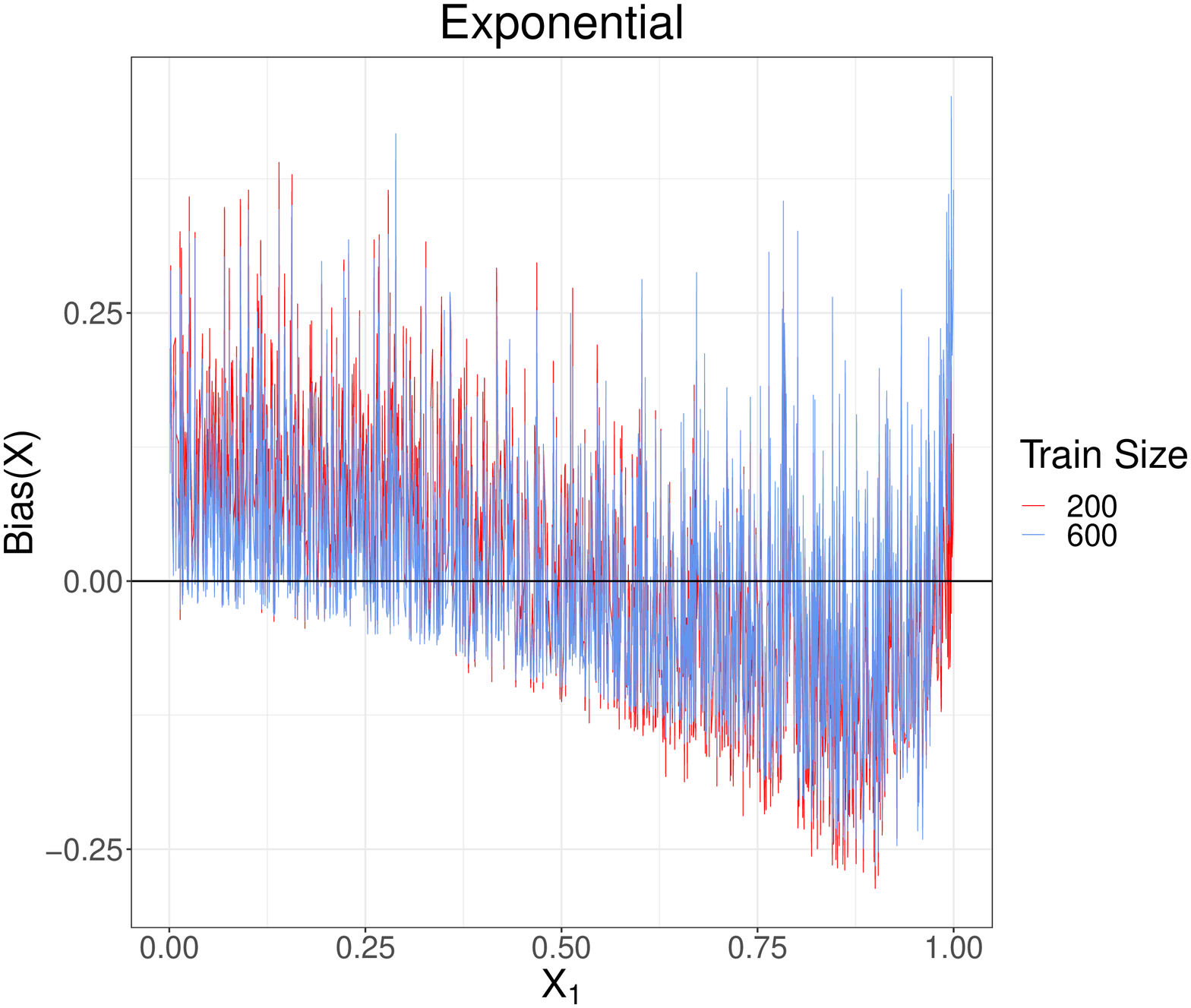}
    \includegraphics[width = 0.496\linewidth]{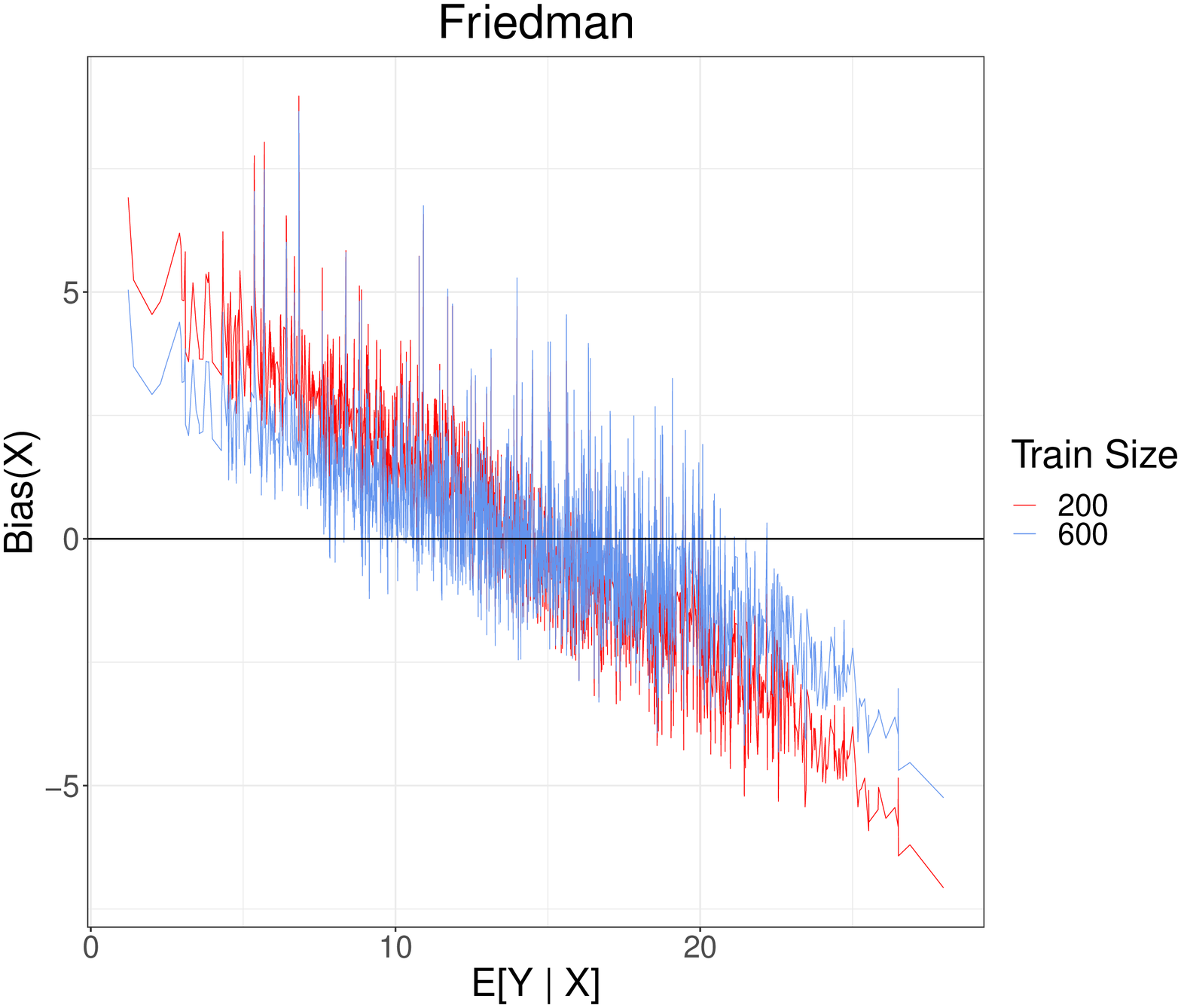}
    \includegraphics[width = 0.496\linewidth]{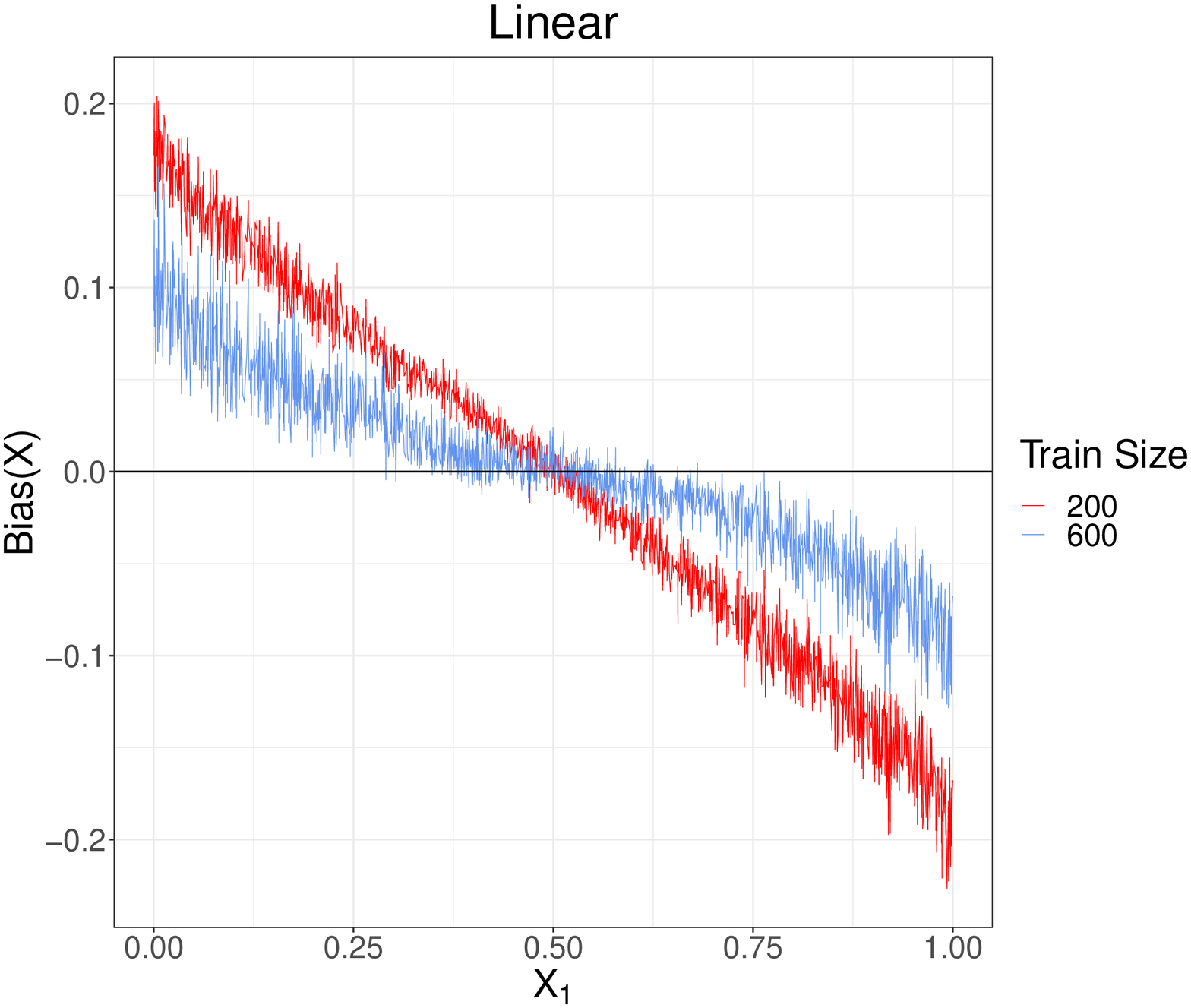}
    \caption{Conditional biases of our stringent version of $\hat{\varphi}^{\text{BC}}(x)$ for the Step, Exponential, Linear, and Friedman datasets (clockwise from top left) over 1,000 simulation repetitions using training sets of 200 units and training sets of 600 units.}
    \label{fig:stringent_bias_behavior}
\end{figure}

\begin{figure}
    \centering
    \includegraphics[width = 0.496\linewidth]{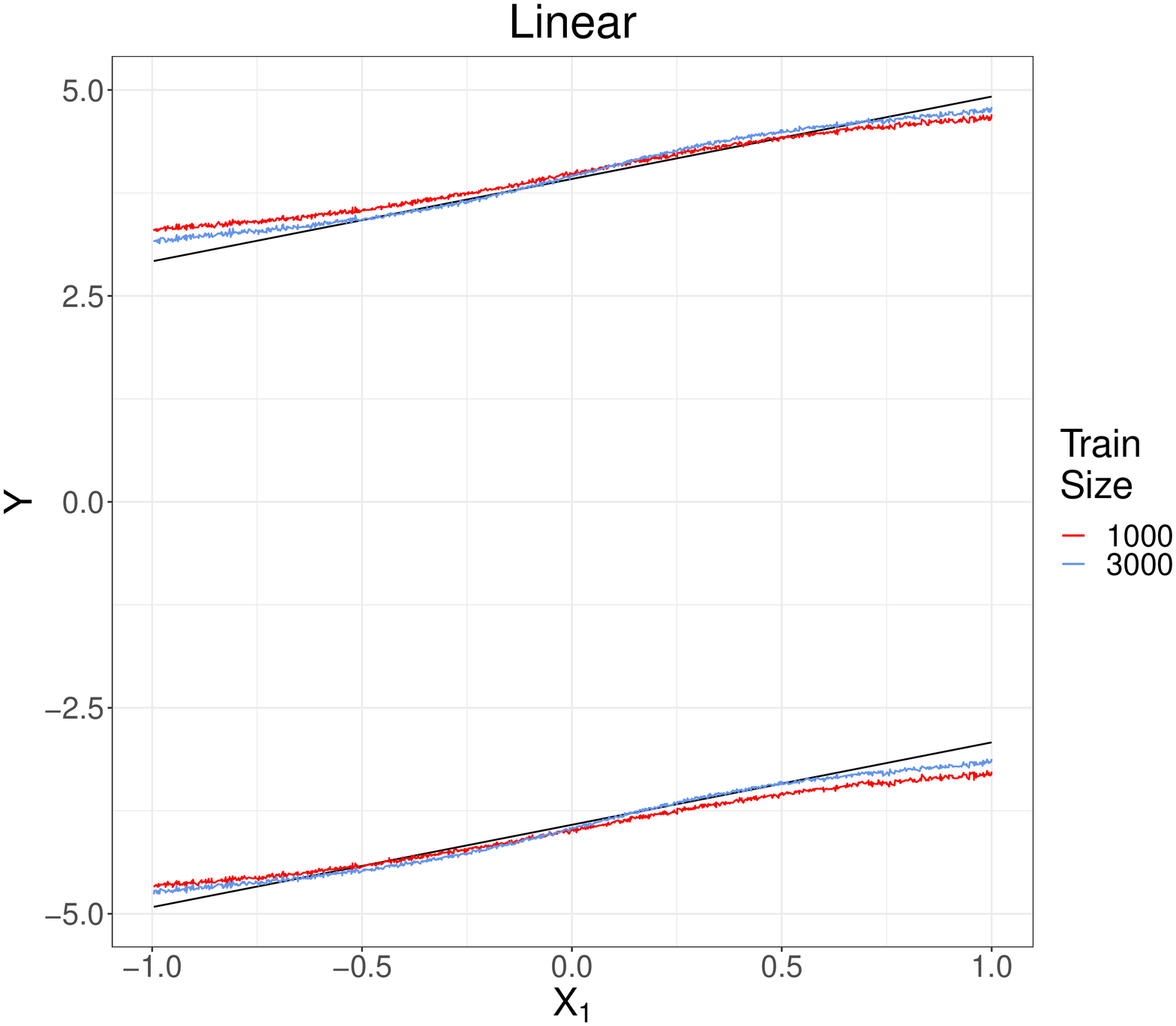}
    \includegraphics[width = 0.496\linewidth]{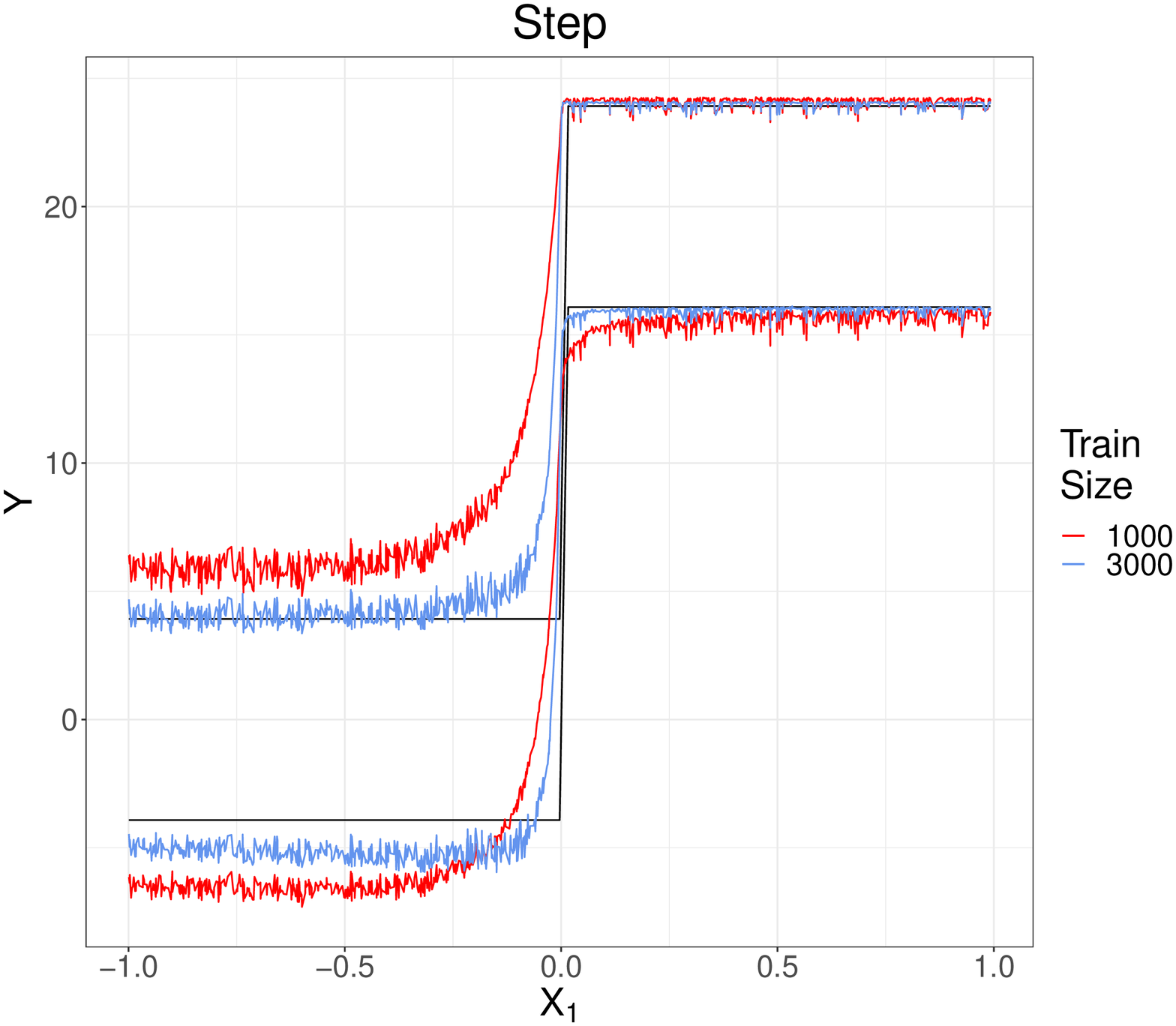}
    \includegraphics[width = 0.496\linewidth]{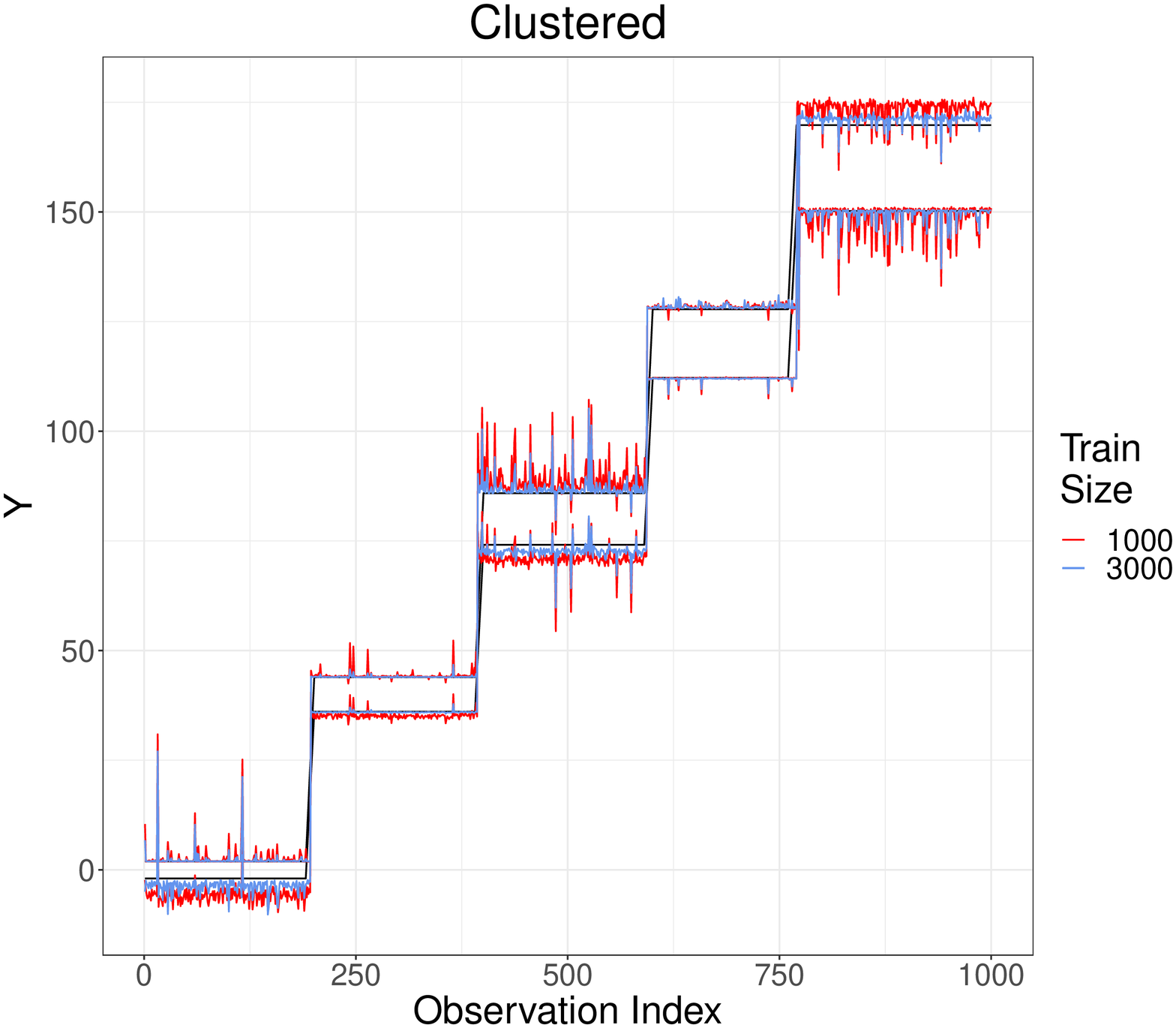}
    \includegraphics[width = 0.496\linewidth]{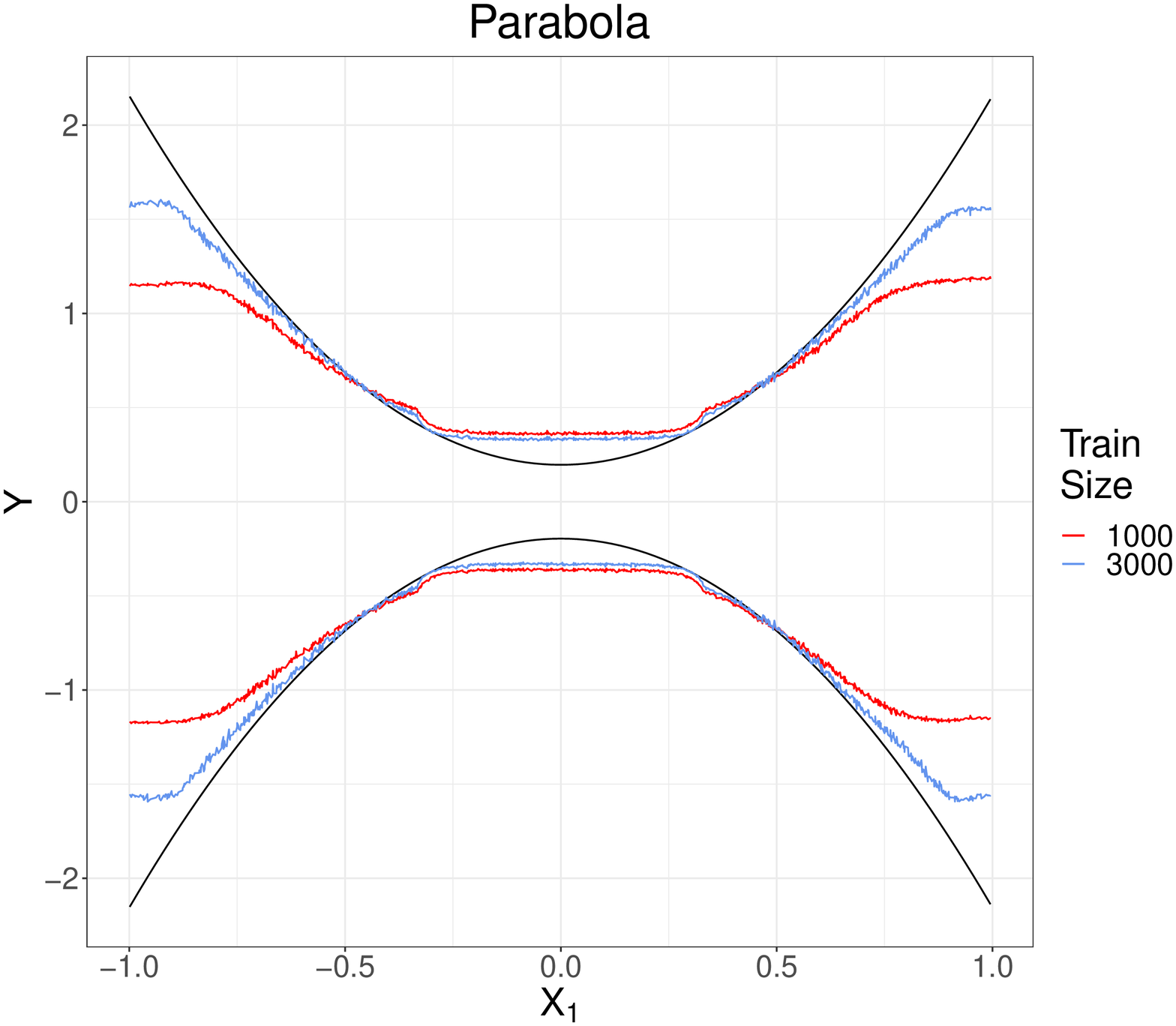}
    \caption{Average upper and lower bounds of 95\% prediction intervals constructed by our stringent version of $\widehat{\text{PI}}_{\alpha}(x)$ for the Linear, Step, Parabola, and Clustered datasets (clockwise from top left) over 1,000 simulation repetitions using training sets of 1,000 units and training sets of 3,000 units. The true target conditional response quantiles are shown in black.}
    \label{fig:stringent_interval_behavior}
\end{figure}

Table \ref{tab:stringent_bias_results} shows the MSB and MSPE of our stringent version of $\hat{\varphi}^{\text{BC}}(x)$, and Figure \ref{fig:stringent_bias_behavior} plots the conditional biases of our stringent version of $\hat{\varphi}^{\text{BC}}(x)$ against the signaling covariate(s). Additionally, Table \ref{tab:stringent_interval_results} shows the coverage rates and widths of our stringent version of $\widehat{\text{PI}}_{\alpha}(x)$, and Figure \ref{fig:stringent_interval_behavior} plots the average conditional response quantiles estimated by our stringent version of $\widehat{\text{PI}}_{\alpha}(x)$ against the true conditional response quantiles for the Linear, Step, Clustered, and Parabola datasets. As expected, our stringent estimator is less data-efficient than our practical estimator due to the sample-splitting, but it behaves similarly to our practical estimator overall. In particular, when given more training units, our stringent estimator performs nearly identically to our practical estimator.

\begin{table}[h]
    \centering
    \begin{tabular}{ccccc}
         \toprule
         & \multicolumn{2}{c}{MSB} & \multicolumn{2}{c}{MSPE} \\
         \cmidrule(r){2-3} \cmidrule(l){4-5}
         Dataset & $\hat{\varphi}^{\text{BC}}(x)$ Original & $\hat{\varphi}^{\text{BC}}(x)$ Rich & $\hat{\varphi}^{\text{BC}}(x)$ Original & $\hat{\varphi}^{\text{BC}}(x)$ Rich\\
         \cline{1-5}
         Baseline      & 0.000 & 0.000 & 1.088  & 1.057\\
         Linear          & 0.010 & 0.002 & 1.099  & 1.062\\
	Step             & 0.702 & 0.227 & 2.154  & 1.402\\
         Exponential & 0.013 & 0.011 & 1.013  & 0.982\\
	Friedman     & 5.040 & 2.711 & 7.327  & 4.446\\
	Boston         &     -     &     -     & 11.178 &      -    \\
	Abalone       &     -     &     -     &  4.844 &      -   \\
	Servo           &     -     &     -     & 24.407 &      -   \\
         \bottomrule
    \end{tabular}
    \caption{Mean squared bias and mean squared prediction error of our stringent version of $\hat{\varphi}^{\text{BC}}(x)$ for each dataset using both the original training set size and, when possible, a richer training set with three times as many units.}
    \label{tab:stringent_bias_results}
\end{table}

\begin{table}[h]
    \centering
    \begin{tabular}{ccc}
         \toprule
         Dataset & $\widehat{\text{PI}}_{\alpha}(x)$ Original & $\widehat{\text{PI}}_{\alpha}(x)$ Rich\\
         \cline{1-3}
	Linear       & 0.946 (7.97)   & 0.948 (7.92)\\
         Clustered & 0.930 (16.57) & 0.942 (13.87)\\
         Step         & 0.946 (8.75)   & 0.945 (8.15)\\
         Friedman & 0.964 (24.67) & 0.968 (21.59)\\
         Parabola  & 0.961 (0.86)   & 0.966 (0.83)\\
         2D           &  0.947 (17.33) & 0.951 (17.15)\\
         Boston     & 0.935 (13.49) & - \\
         Abalone   & 0.940 (8.25)   & - \\
	Servo       & 0.906 (22.77) & - \\
         \bottomrule
    \end{tabular}
    \caption{Average coverage rates and widths of 95\% prediction intervals constructed by our stringent version of $\widehat{\text{PI}}_{\alpha}(x)$ using both the original training set size and, when possible, a richer training set with three times as many units.}
    \label{tab:stringent_interval_results}
\end{table}

\subsection{Implementation Details and Additional Results for $\widehat{\text{PI}}_{\alpha}^{\text{GRF}}(x)$}
\label{ap:grf}

As discussed in Section \ref{sec:interval_sim}, we also adapted our method of prediction interval estimation to the quantile regression variant of generalized random forests, with minor changes. In particular, we fit a generalized random forest to the training set using the parameters identified in Appendix \ref{ap:params}. We then computed the out-of-bag prediction errors as the difference between each training observation's observed response and the generalized random forest's out-of-bag prediction of its median response. Next, for a given test observation with covariates $x$, we computed the weight of each training observation by counting the number of trees in which the training observation was both a cohabitant of $x$ and part of the honest subsample that was used to populate the tree's nodes but not to determine its splits; we normalized the weights to sum to one.

\begin{figure}
    \centering
    \includegraphics[width = 0.496\linewidth]{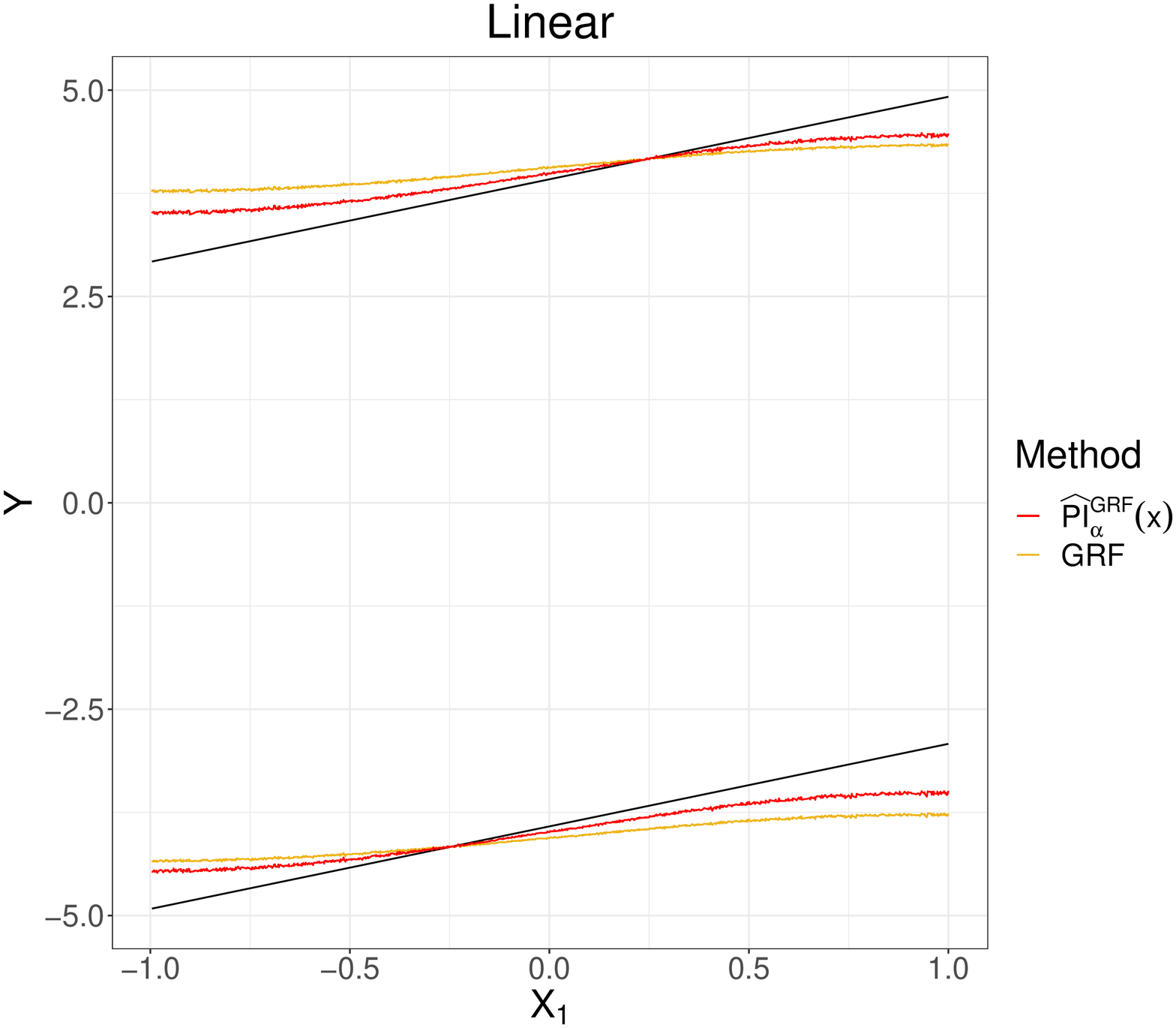}
    \includegraphics[width = 0.496\linewidth]{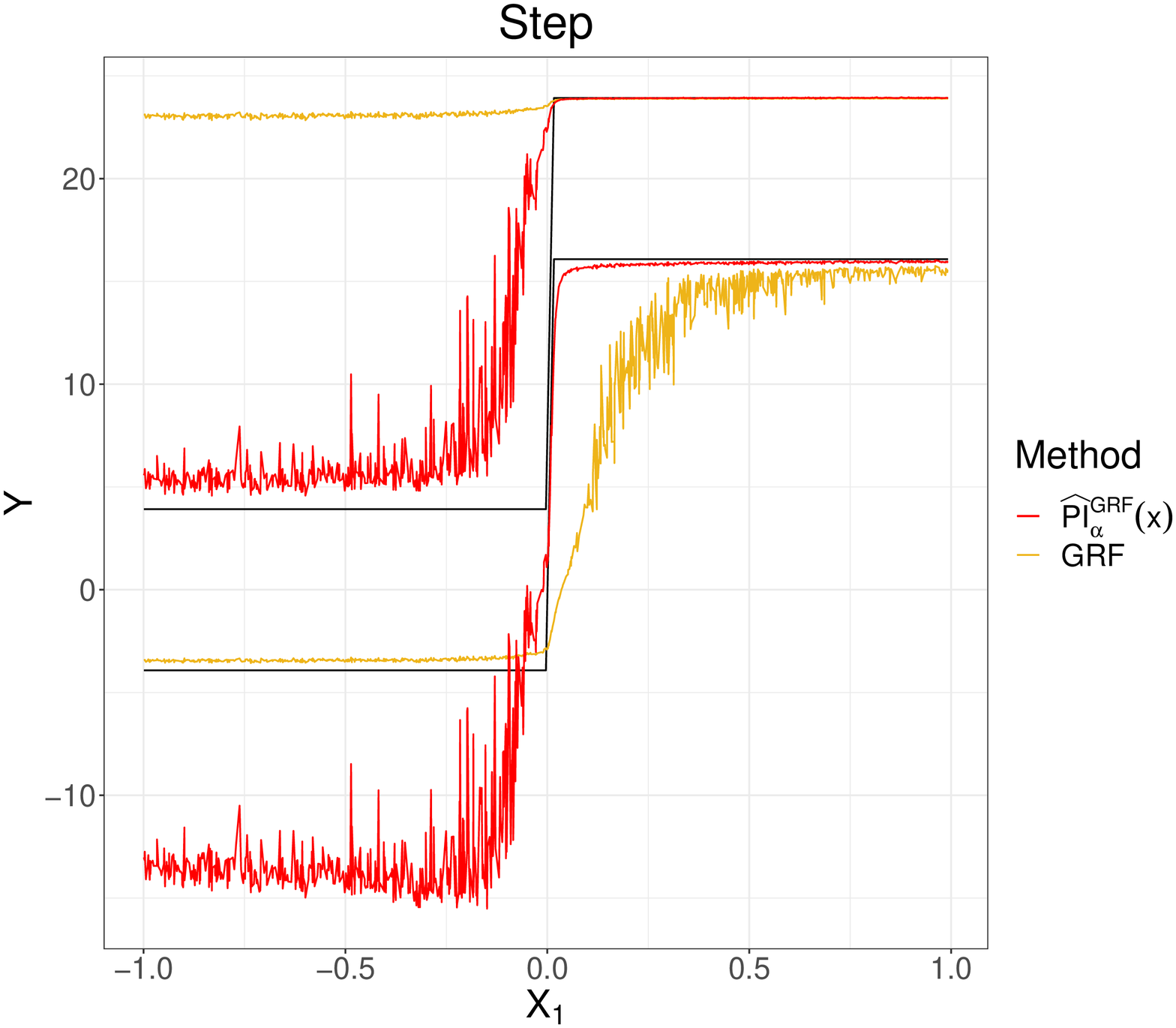}
    \includegraphics[width = 0.496\linewidth]{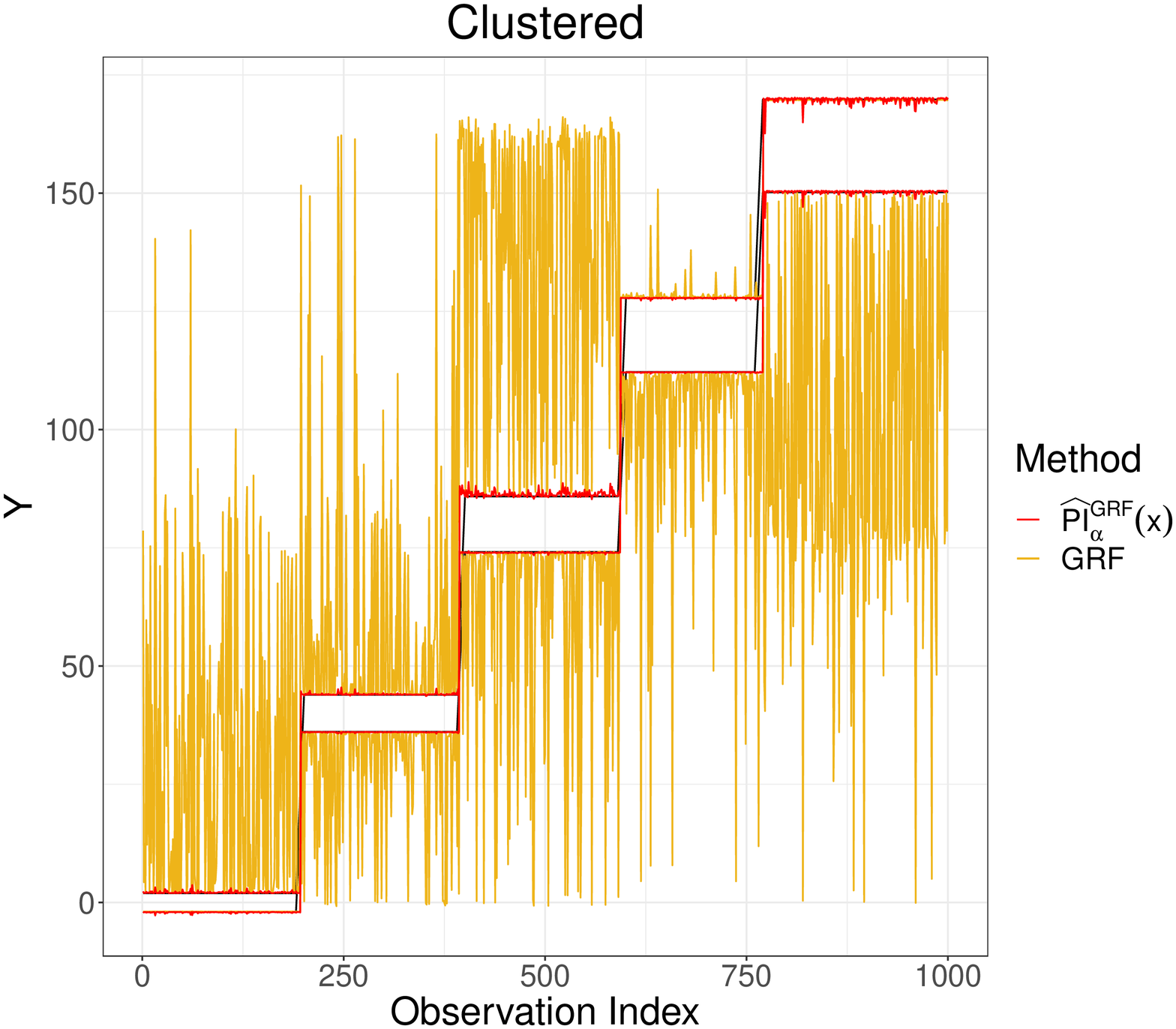}
    \includegraphics[width = 0.496\linewidth]{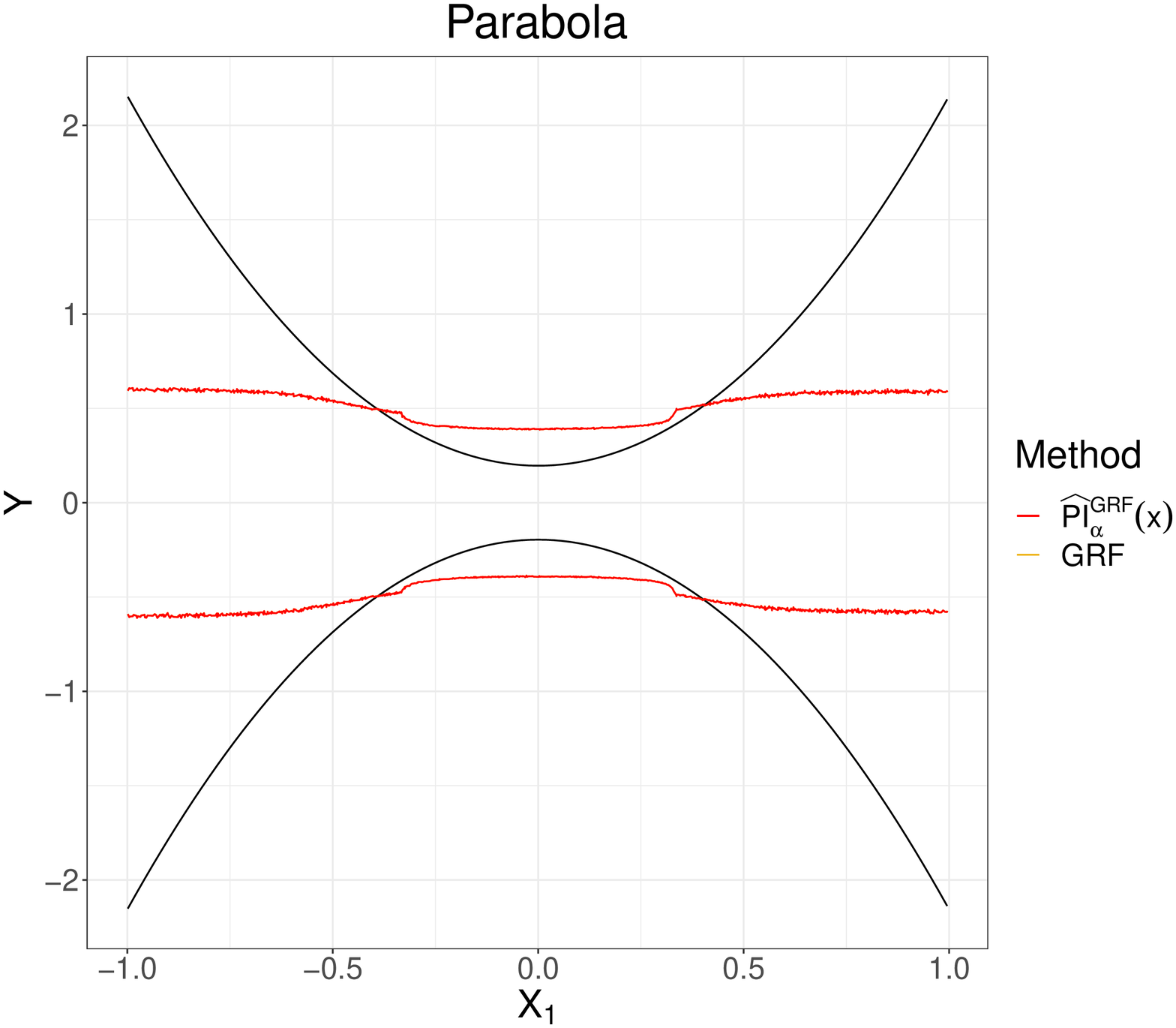}
    \caption{Average upper and lower bounds of 95\% prediction intervals constructed by $\widehat{\text{PI}}_{\alpha}^{\text{GRF}}(x)$ for the Linear, Step, Parabola, and Clustered datasets (clockwise from top left) over 1,000 simulation repetitions. The true target conditional response quantiles are shown in black. The average upper and lower bounds of 95\% prediction intervals constructed by generalized random forests are reproduced from Figure \ref{fig:interval_behavior} for ease of comparison.}
    \label{fig:grf_interval_behavior}
\end{figure}

Table \ref{tab:grf_interval_results} shows the average coverage rates and widths of our adaptation to generalized random forests, which we denote by $\widehat{\text{PI}}_{\alpha}^{\text{GRF}}(x)$. We also reproduced the corresponding results for generalized random forests from Table \ref{tab:interval_results} for ease of comparison. Figure \ref{fig:grf_interval_behavior} plots the average conditional response quantiles estimated by $\widehat{\text{PI}}_{\alpha}^{\text{GRF}}(x)$ against the true conditional response quantiles for the Linear, Step, Clustered, and Parabola datasets. Again, we also reproduced the corresponding results for generalized random forests from Figure \ref{fig:interval_behavior} for ease of comparison. Notably, $\widehat{\text{PI}}_{\alpha}^{\text{GRF}}(x)$ generally produced better-calibrated prediction intervals, with coverage rates closer to the desired 95\% rate. Additionally, $\widehat{\text{PI}}_{\alpha}^{\text{GRF}}(x)$ generally produced conditional quantile estimates that qualitatively behaved more like the true conditional quantiles across the covariate space.

\begin{table}[h]
    \centering
    \begin{tabular}{ccc}
         \toprule
         Dataset & $\widehat{\text{PI}}_{\alpha}^{\text{GRF}}(x)$ & GRF\\
         \cline{1-3}
	Linear       & 0.950 (7.97)    &   0.952 (8.11)\\
         Clustered & 0.947 (11.94)   &  0.966 (41.27)\\
         Step         & 0.951 (8.82)    &   0.962 (12.10)\\
         Friedman & 0.979 (25.66)  &   0.991 (45.50)\\
         Parabola  & 0.960 (0.84)    &   0.960 (0.84)\\
         2D           &  0.952 (17.34)  &   0.962 (18.87)\\
         Boston     & 0.965 (15.54)   &   0.994 (23.92)\\
         Abalone   & 0.975 (8.49)     &   0.982 (9.21)\\
	Servo       & 0.968 (24.43)   &   0.985 (37.29)\\
         \bottomrule
    \end{tabular}
    \caption{Average coverage rates and widths of 95\% prediction intervals constructed by $\widehat{\text{PI}}_{\alpha}^{\text{GRF}}(x)$. The average coverage rates and widths of 95\% prediction intervals constructed by generalized random forests are reproduced from Table \ref{tab:interval_results} for ease of comparison.}
    \label{tab:grf_interval_results}
\end{table}

\newpage

\section{Proofs of Proposition \ref{prop:errorConverge} and Theorem \ref{th:main}}
\label{ap:proof}

\begin{proof}[Proof of Proposition \ref{prop:errorConverge}]
Fix $x \in \mathcal{X}$. Since $\theta_1, \ldots, \theta_B$ are i.i.d. and independent of $\mathcal{D}_n$, the weak law of large numbers implies that, conditionally on $\mathcal{D}_n$,
\[
\frac{1}{B}\sum_{b = 1}^{B}\sum_{j = 1}^{n}w_j(x, \theta_b)Y_j \toProb \mu := \mathbb{E}\left[\sum_{j = 1}^{n}w_j(x, \theta)Y_j \mid \mathcal{D}_n\right],\hspace{20pt}B \to \infty,
\]
where the expectation is taken over $\theta$. Then, by Slutsky's Theorem, $E \mid x \toDist \mu - Y \mid x$. An analogous argument implies that $E^* \mid x \toDist \mu - Y \mid x$ as well. Recall that (pointwise) convergence in distribution implies uniform convergence for continuous cumulative distribution functions. Since $F_{Y}(\cdot \mid x)$ is continuous by assumption, $F_{\mu - Y}(\cdot \mid x)$ is continuous as well. Therefore,
\begin{align*}
\lim_{B \to \infty}\sup_{e \in \mathbb{R}}&\left|F_{E^*}(e \mid x) - F_{E}(e \mid x)\right|\\
&= \lim_{B \to \infty}\sup_{e \in \mathbb{R}}\left|(F_{E^*}(e \mid x) - F_{\mu - Y}(e \mid x)) + (F_{E}(e \mid x) - F_{\mu - Y}(e \mid x))\right|\\
&\leq \lim_{B \to \infty}\sup_{e \in \mathbb{R}}\left|F_{E^*}(e \mid x) - F_{\mu - Y}(e \mid x)\right| + \left|F_{E}(e \mid x) - F_{\mu - Y}(e \mid x)\right|\\
&\leq \lim_{B \to \infty}\sup_{e \in \mathbb{R}}\left|F_{E^*}(e \mid x) - F_{\mu - Y}(e \mid x)\right| + \lim_{B \to \infty}\sup_{e \in \mathbb{R}}\left|F_{E}(e \mid x) - F_{\mu - Y}(e \mid x)\right|\\
&= 0,
\end{align*} 
which completes the proof.
\end{proof}\\[2mm]

\noindent Before proving Theorem \ref{th:main}, we establish additional notation. First, let $\Omega_v$ denote the set of observations---$\mathcal{J} \cup \mathcal{K}$---and parameters that fully define the second random forest (Step 3 of Section \ref{sec:alg}). Second, let $M_n$ denote the maximum possible value of $\vx{i}$, which is decreasing in $n$ by Assumption \ref{as:propobs}. Finally, for any $\delta > 0$, let $\gamma_n := \Pr(\neg\MM)$ denote the probability that $\delta$-stability is not realized for the first random forest's prediction of the $i^{\text{th}}$ training unit in $\mathcal{K}$, which is decreasing in $n$ by Assumption \ref{as:consistency}.\\

\begin{proof}[Proof of Theorem \ref{th:main}]
Fix $x \in [0, 1]^p$. Let the random variables $U_i, i = 1, \ldots, n$, be defined as the quantiles of $E_i$  given $X_i$:
\[
U_i := F_E(E_i \mid X_i).
\]
Notice that, since $E_i$ follows the distribution of $E \mid X_i$, $U_i \sim \text{Unif}[0, 1]$. Additionally, Assumption \ref{as:monotone} implies that the event $\{E_i \leq e\}$ is equivalent to the event $\{U_i \leq F_E(e \mid X_i)\}$. Using this equivalence, we have
\begin{align*}
\hat{F}_E(e \mid x) &= \sum_{i = 1}^{n}v_i(x)\mathbbm{1}(E_i \leq e)\\
&= \sum_{i = 1}^{n}v_i(x)\mathbbm{1}(U_i \leq F_E(e \mid X_i))\\
&= \sum_{i = 1}^{n}v_i(x)\mathbbm{1}(U_i \leq F_E(e \mid x)) + \sum_{i = 1}^{n}v_i(x)\big(\mathbbm{1}(U_i \leq F_E(e \mid X_i)) - \mathbbm{1}(U_i \leq F_E(e \mid x))\big),
\end{align*}
so $\left|\hat{F}_E(e \mid x) - F_E(e \mid x)\right|$ is bounded above by
\begin{align}
\left|\hat{F}_E(e \mid x) - F_E(e \mid x)\right| \leq &\underbrace{\left|\sum_{i = 1}^{n}v_i(x)\mathbbm{1}(U_i \leq F_E(e \mid x)) -  F_E(e \mid x)\right|}_{\text{``variance term''}}\nonumber\\
&+\underbrace{\left|\sum_{i = 1}^{n}v_i(x)\big(\mathbbm{1}(U_i \leq F_E(e \mid X_i)) - \mathbbm{1}(U_i \leq F_E(e \mid x))\big)\right|}_{\text{``shift term''}}. \label{eq:mainprob}
\end{align}
As mentioned in \citet{meinshausen2006}, the first term on the right side of (\ref{eq:mainprob}) can be thought of as a variance-type term, while the second term can be thought of as reflecting the shift in the underlying error distribution across the covariate space. In the next two subsections, we show that each term converges to zero in probability.

\subsection*{Bounding the Variance Term}
\label{sec:var}
Taking the supremum over $e$ of the variance term yields
\[
\sup_{e \in \mathbb{R}}\left|\sum_{i = 1}^{n}v_i(x)\mathbbm{1}(U_i \leq F_E(e \mid x)) -  F_E(e \mid x)\right| = \sup_{z \in [0, 1]}\left|\sum_{i = 1}^{n}v_i(x)\mathbbm{1}(U_i \leq z) - z\right|.
\]
It suffices to prove that, for all $z \in [0, 1]$,
\begin{equation}
\label{eq:op1}
\left|\sum_{i = 1}^{n}v_i(x)\mathbbm{1}(U_i \leq z) - z\right| = o_p(1).
\end{equation}
Since the forest used for the weights $v_i(x)$ is built on a separate subset of the training data from the forest used for the predictions $\hat{\varphi}(X_i)$, and the prediction of the $i^{\text{th}}$ observation in $\mathcal{K}$ does not depend on the other $n - 1$ observations in $\mathcal{K}$, conditioning on $X_i$ yields sufficient independence to evaluate the expectation of the weighted average inside (\ref{eq:op1}):
\begin{align}
\mathbb{E}\sum_{i = 1}^{n}v_i(x)\mathbbm{1}(U_i \leq z) &= \sum_{i = 1}^{n}\mathbb{E}\big[\mathbb{E}[v_i(x)\mathbbm{1}(U_i \leq z) \mid X_i]\big] \nonumber\\
&= \sum_{i = 1}^{n}\mathbb{E}\big[\mathbb{E}[v_i(x) \mid X_i]\Pr(U_i \leq z \mid X_i)\big] \nonumber\\
&= z\mathbb{E}\sum_{i = 1}^{n}v_i(x) \nonumber\\
&= z.\label{eq:expectedvalue}
\end{align}
Moreover, since the variance of a summation is equal to the summation of the covariances,
\begin{align*}
\Var\bigg(\sum_{i = 1}^{n}\vx{i}\ux{i}\bigg) = &\sum_{i = 1}^{n}\Var(\vx{i}\ux{i})\\
&+ \sum_{i \neq j}\text{Cov}(\vx{i}\ux{i}, \vx{j}\ux{j}),
\end{align*}
with each summation converging to zero by Lemmas \ref{lem:vartermsumofvar} and \ref{lem:vartermsumofcov}.

\subsection*{Bounding the Shift Term}

Next, we show that the shift term converges to zero in probability; again, it suffices to show convergence for all $e \in \mathbb{R}$. As an intermediate result, we first show that
\begin{equation}
\label{eq:shift472}
\sum_{i = 1}^{n}v_i(x)\big(\mathbbm{1}(U_i \leq F_E(e \mid X_i)) - \mathbbm{1}(U_i \leq F_E(e \mid x))\big) \toProb \sum_{i = 1}^{n}\vx{i}\big(F_E(e \mid X_i) - F_E(e \mid x)\big).
\end{equation}
By the triangle inequality, the union bound, and (\ref{eq:op1}), we can reduce the task of showing (\ref{eq:shift472}) to simply showing that
\begin{equation}
\label{eq:shift473}
\sum_{i = 1}^{n}\vx{i}\mathbbm{1}(U_i \leq F_E(e \mid X_i)) - \sum_{i = 1}^{n}\vx{i}F_E(e \mid X_i) \toProb 0.
\end{equation}
We do so by showing that the left side of (\ref{eq:shift473}) has expectation zero and decreasing variance. Since $\vx{i}$ and $\mathbbm{1}(U_i \leq F_E(e \mid X_i))$ are independent conditional on $X_i$, and $\Pr(U_i \leq F_E(e \mid X_i) \mid X_i) = F_E(e \mid X_i)$, a direct application of the tower property conditioning on $X_i$ yields the identity
\begin{equation}
\label{eq:tower1}
\mathbb{E}\left[\vx{i}\mathbbm{1}(U_i \leq F_E(e \mid X_i))\right] = \mathbb{E}\left[\vx{i}F_E(e \mid X_i)\right].
\end{equation}
Thus, (\ref{eq:tower1}) and the linearity of expectation implies that
\[
\mathbb{E}\left[\sum_{i = 1}^{n}\vx{i}\mathbbm{1}(U_i \leq F_E(e \mid X_i)) - \sum_{i = 1}^{n}\vx{i}F_E(e \mid X_i)\right] = 0.
\]
Next, we again decompose the variance of the summation into the sum of covariances:
\begin{align}
\Var&\left(\sum_{i = 1}^{n}\vx{i}\mathbbm{1}(U_i \leq F_E(e \mid X_i)) - \sum_{i = 1}^{n}\vx{i}F_E(e \mid X_i)\right) \nonumber\\
&= \sum_{i = 1}^{n}\Var\left(\vx{i}\left(\mathbbm{1}(U_i \leq F_E(e \mid X_i)) - F_E(e \mid X_i)\right)\right) \nonumber\\
&\hphantom{{}=}+ \sum_{i \neq j}\text{Cov}\Big(\vx{i}\big(\mathbbm{1}(U_i \leq F_E(e \mid X_i)) - F_E(e \mid X_i)\big), \nonumber\\
&\hphantom{{}=+ \sum_{i \neq j}\text{Cov}\Big(}\vx{j}\big(\mathbbm{1}(U_j \leq F_E(e \mid X_j)) - F_E(e \mid X_j)\big)\Big),\label{eq:part2vardecompose}
\end{align}
with each summation converging to zero by Lemmas \ref{lem:shifttermsumofvar} and \ref{lem:shifttermsumofcov}. With our intermediate result (\ref{eq:shift472}) complete, we note that, by Lipschitz continuity of the conditional prediction error distribution (Assumption \ref{as:error}), it only remains to be shown that
\begin{equation}
\label{eq:final}
\sum_{i = 1}^{n}\vx{i}\left\|X_i - x\right\|_1 = o_p(1).
\end{equation}
This follows from Lemma 2 of \citet{meinshausen2006}. In particular, recall that
\[
v_i(x) = \lim_{B \to \infty}\frac{1}{B}\sum_{b = 1}^{B}\frac{\#\{Z_i \in \mathcal{D}^*_{n,b}\}\mathbbm{1}(X_i \in R_{\ell(x, \theta_b)})}{\sum_{j = 1}^{n}\#\{Z_j \in \mathcal{D}^*_{n,b}\}\mathbbm{1}(X_j \in R_{\ell(x, \theta_b)})}
\]
so showing (\ref{eq:final}) is equivalent to showing that
\[
\lim_{B \to \infty}\frac{1}{B}\sum_{b = 1}^{B}\sum_{i = 1}^{n}\frac{\#\{Z_i \in \mathcal{D}^*_{n,b}\}\mathbbm{1}(X_i \in R_{\ell(x, \theta_b)})}{\sum_{j = 1}^{n}\#\{Z_j \in \mathcal{D}^*_{n,b}\}\mathbbm{1}(X_j \in R_{\ell(x, \theta_b)})}\|X_i - x\|_1 \toProb 0.
\]
Therefore, it suffices to show that, for a single tree,
\[
\sum_{i = 1}^{n}\frac{\#\{Z_i \in \mathcal{D}^*_n\}\mathbbm{1}(X_i \in R_{\ell(x, \theta)})}{\sum_{j = 1}^{n}\#\{Z_j \in \mathcal{D}^*_n\}\mathbbm{1}(X_j \in R_{\ell(x, \theta)})}\|X_i - x\|_1 \toProb 0.
\]
Following the argument in the proof of Theorem 1 of \citet{meinshausen2006}, we can decompose the rectangular subspace $R_{\ell(x, \theta)} \subseteq [0, 1]^p$ of leaf $\ell(x, \theta)$ of the tree into the intervals $I(x, m, \theta) \subseteq [0, 1]$ for $m = 1, \ldots, p$:
\[
R_{\ell(x, \theta)} = \otimes_{m = 1}^{p}I(x, m, \theta).
\]
Note that $X_i \notin I(x, m, \theta)$ implies that $\mathbbm{1}(X_i \in R_{\ell(x,\theta)}) = 0$. Thus, it suffices to show that $\max_{m}\left|I(x, m, \theta)\right| = o_{p}(1)$, which Lemma 2 of \citet{meinshausen2006} accomplishes.
\end{proof}\\[2mm]

\noindent Before stating and proving Lemmas \ref{lem:vartermsumofvar}-\ref{lem:shifttermsumofcov}, we establish Lemma \ref{lem:asymptotics} for use in Lemmas \ref{lem:vartermsumofcov} and \ref{lem:shifttermsumofcov}.

\begin{lemma2}
\label{lem:asymptotics}
Under Assumptions \ref{as:uniform}-\ref{as:weight}, we have the following asymptotic results for any $\delta \in \left(0, \delta_0\right)$:
\begin{align*}
\gamma_n\sum_{i = 1}^{n}\EE\left[\vx{i} \mid \MM\right] &\to 0,\hspace{20pt}n \to \infty;\\
\gamma_n\sum_{i = 1}^{n}\EE\left[\vx{i} \mid \neg\MM\right] &\to 0,\hspace{20pt}n \to \infty;\text{ and}\\
\sum_{i = 1}^{n}\EE\left[\vx{i}^2 \mid \MM\right] &\to 0,\hspace{20pt}n \to \infty.
\end{align*}
\end{lemma2}

\begin{proof}[Proof of Lemma \ref{lem:asymptotics}]
Fix $\delta \in \left(0, \delta_0\right)$ and let $\epsilon > 0$. Assumption \ref{as:weight} implies that there exists a constant $c > 0$ and $N_1 \in \mathbb{N}$ so that, for $n \geq N_1$, $\EE\left[\vx{i} \mid \MM\right] \leq c / n$ and $\EE\left[\vx{i} \mid \neg\MM\right] \leq c / n.$ Fix that value of $c$. Since the random forest is stable by Assumption \ref{as:consistency}, there exists $N_2 \in \mathbb{N}$ so that, for $n \geq N_2$, $\gamma_n < \epsilon / c.$ Moreover, by Assumption \ref{as:propobs}, the minimum number of observations in each node is growing, so the maximum possible weight $M_n$ given to any one unit is decreasing in $n$. Thus, there exists $N_3 \in \mathbb{N}$ so that, for $n \geq N_3$, $M_n < \epsilon / c$. Therefore, for $n \geq \max\{N_1, N_2, N_3\}$,
\begin{align*}
\gamma_n\sum_{i = 1}^{n}\EE\left[\vx{i} \mid \MM\right] &< \frac{\epsilon}{c}\sum_{i = 1}^{n}\frac{c}{n} = \epsilon,\\
\gamma_n\sum_{i = 1}^{n}\EE\left[\vx{i} \mid \neg\MM\right] &< \frac{\epsilon}{c}\sum_{i = 1}^{n}\frac{c}{n} = \epsilon,\text{ and}\\
\sum_{i = 1}^{n}\EE\left[\vx{i}^2 \mid \MM\right] &\leq M_n\sum_{i = 1}^{n}\EE\left[\vx{i} \mid \MM\right] < \frac{\epsilon}{c}\sum_{i = 1}^{n}\frac{c}{n} = \epsilon.
\end{align*}
This completes the proof.
\end{proof}

\begin{lemma2}
\label{lem:vartermsumofvar}
Under Assumptions \ref{as:uniform}-\ref{as:weight},
\[
\sum_{i = 1}^{n}\Var(\vx{i}\ux{i}) \to 0,\hspace{20pt}n \to \infty.
\]
\end{lemma2}
\begin{proof}[Proof of Lemma \ref{lem:vartermsumofvar}]
By the law of total variance,
\begin{align}
\sum_{i = 1}^{n}&\Var(\vx{i}\ux{i}) \nonumber\\
&= \sum_{i = 1}^{n}\Var(\EE[\vx{i}\ux{i} \mid \Omega_v \setminus \{Y_i\}]) + \EE[\Var(\vx{i}\ux{i} \mid \Omega_v \setminus \{Y_i\})] \nonumber\\
&= \sum_{i = 1}^{n}\Var(\vx{i}\Pr(U_i \leq z \mid X_i)) + \EE[\vx{i}^2\Var(\ux{i} \mid X_i)] \nonumber\\
&= \sum_{i = 1}^{n}z^2\Var(\vx{i}) + z(1 - z)\EE[\vx{i}^2] \nonumber\\
&\leq \sum_{i = 1}^{n}\Var(\vx{i}) + \EE[\vx{i}^2]. \label{eq:varbreak}
\end{align}
Notice that
\begin{equation}
\label{eq:minweight}
\sum_{i = 1}^{n}\Var(\vx{i}) \leq \sum_{i = 1}^{n}\EE[\vx{i}^2] \leq M_n\EE\sum_{i = 1}^{n}\vx{i} = M_n.
\end{equation}
Since the minimum number of observations in each node is growing by Assumption \ref{as:propobs}, the maximum possible weight given to any observation is decreasing in $n$---that is, $M_n \to 0$. Thus, plugging the bound given by (\ref{eq:minweight}) into (\ref{eq:varbreak}) yields the desired result:
\[
\lim_{n \to \infty}\sum_{i = 1}^{n}\Var(\vx{i}\ux{i}) \leq \lim_{n \to \infty}2M_n = 0.
\]
This completes the proof.
\end{proof}\\[2mm]

\begin{lemma2}
\label{lem:vartermsumofcov}
Under Assumptions \ref{as:uniform}-\ref{as:weight},
\[
\sum_{i \neq j}\text{Cov}(\vx{i}\ux{i}, \vx{j}\ux{j}) \to 0,\hspace{20pt}n \to \infty.
\]
\end{lemma2}
\begin{proof}[Proof of Lemma \ref{lem:vartermsumofcov}]
Since
\begin{align*}
\sum_{i \neq j}\text{Cov}&(\vx{i}\ux{i}, \vx{j}\ux{j})\\
&= \sum_{i \neq j}\EE[\vx{i}\vx{j}\ux{i}\ux{j}] - \EE[\vx{i}\ux{i}]\EE[\vx{j}\ux{j}]\\
&\to \sum_{i \neq j}\EE[\vx{i}\vx{j}\ux{i}\ux{j}] - z^2,
\end{align*}
it suffices to show that
\[
\lim_{n \to \infty}\left|\sum_{i \neq j}\EE[\vx{i}\vx{j}\ux{i}\ux{j}] - z^2\right| = 0.
\]
Let $\epsilon > 0$. By uniform continuity of the conditional response distribution, there exists $\delta_1 > 0$ so that
\begin{equation}
\label{eq:continuity}
\left|y_1 - y_2\right| < 2\delta_1 \implies \left|F_{Y}(y_1 \mid x) - F_{Y}(y_2 \mid x)\right| < \epsilon / 3.
\end{equation}
Fix $\delta < \min\left\{\delta_0, \delta_1\right\}$. Then, by Lemma \ref{lem:asymptotics}, there exists $N \in \mathbb{N}$ so that, for $n \geq N$, terms identified in Lemma \ref{lem:asymptotics} that appear in this proof sum to at most $\epsilon / 3$; for concision, we note these leftover terms where they appear and then cite Lemma \ref{lem:asymptotics} to drop them.

\subsection*{Independence of the Error Terms Conditional on Realized $\delta$-Stability}

We use the law of total expectation to condition on the realization of $\delta$-stability of the random forest prediction of the $i^{\text{th}}$ training observation, then apply the triangle inequality, noting that leftover terms converge to zero by Lemma \ref{lem:asymptotics}, to bound
\[
\left|\sum_{i \neq j}\EE[\vx{i}\vx{j}\ux{i}\ux{j}] - z^2\right|
\]
above by
\begin{equation}
\label{eq:workbloh}
\left|\sum_{i \neq j}\EE[\vx{i}\vx{j}\ux{i}\ux{j} \mid \MM] - z^2\right|,
\end{equation}
discounting leftover terms. Next, we use the realized $\delta$-stability of the $i^{\text{th}}$ prediction to achieve independence of the $i^{\text{th}}$ and $j^{\text{th}}$ error terms, then eliminate the $j^{\text{th}}$ error term. Without loss of generality of whether the $\delta$ is added or subtracted, we can bound (\ref{eq:workbloh}) by substituting in the bound on $\hat{\varphi}(X_i)$ implied by $\MM$, then use the tower property conditioning on the $j^{\text{th}}$ covariate:
\begin{align}
&\left|\sum_{i \neq j}\EE[\vx{i}\vx{j}\ux{i}\ux{j} \mid \MM] - z^2\right| \nonumber\\
&\hphantom{===}\leq \left|\sum_{i \neq j}\EE[\vx{i}\vx{j}\mathbbm{1}(\varphi(X_i) - \delta - Y_i \leq F^{-1}_E(z \mid X_i))\ux{j} \mid \MM] - z^2\right| \nonumber\\
&\hphantom{===}= \Bigg|\sum_{i \neq j}\EE[\EE[\vx{i}\vx{j}\mathbbm{1}(\varphi(X_i) - \delta - Y_i \leq F^{-1}_E(z \mid X_i)) \mid X_j, \MM] \nonumber\\
&\hphantom{{}==== \Bigg|\sum_{i \neq j}\EE[}\cdot\Pr(U_j \leq z \mid X_j, \MM) \mid \MM] - z^2\Bigg|. \label{eq:worktwo}
\end{align}
We evaluate the conditional probability in (\ref{eq:worktwo}) by exploiting the fact that $\delta$-stability of the $i^{\text{th}}$ prediction is independent of $X_j$, so $\Pr(\MM \mid X_j) = \Pr(\MM) = 1 - \gamma_n$. This, along with the law of total probability, the triangle inequality, and Assumption \ref{as:consistency}, implies that 
\begin{align}
\left|\Pr(U_j \leq z \mid X_j, \MM) - z\right| &= \left|\frac{\Pr(U_j \leq z \mid X_j) - \Pr(U_j \leq z \mid X_j, \neg\MM)\gamma_n}{1 - \gamma_n} - z\right| \nonumber\\
&\leq \frac{2\gamma_n}{1 - \gamma_n}.\label{eq:zxcv}
\end{align}
Moreover, Lemma \ref{lem:asymptotics} implies that
\begin{equation}
\label{eq:poiu}
\frac{2\gamma_n}{1 - \gamma_n}\sum_{i \neq j}\EE\left[\vx{i}\vx{j}\mathbbm{1}(\varphi(X_i) - \delta - Y_i \leq F^{-1}_E(z \mid X_i)) \mid \MM\right] \to 0.
\end{equation}
We therefore substitute into (\ref{eq:worktwo}) our upper bound on $\Pr(U_j \leq z \mid X_j, \MM)$ given by (\ref{eq:zxcv}) via the triangle inequality, then apply (\ref{eq:poiu}) to eliminate the leftover term, ultimately bounding (\ref{eq:worktwo}) above by
\begin{equation}
\label{eq:subvar}
z\left|\sum_{i \neq j}\EE[\vx{i}\vx{j}\mathbbm{1}(\varphi(X_i) - \delta - Y_i \leq F^{-1}_E(z \mid X_i)) \mid \MM] - z\right|,
\end{equation}
discounting leftover terms. Since $z \leq 1$, we drop the $z$ outside of the absolute value in (\ref{eq:subvar}). Next, we eliminate $\vx{j}$ terms by applying the triangle inequality and Lemma \ref{lem:asymptotics} as follows:
\begin{align}
&\left|\sum_{i \neq j}\EE[\vx{i}\vx{j}\mathbbm{1}(\varphi(X_i) - \delta - Y_i \leq F^{-1}_E(z \mid X_i)) \mid \MM] - z\right| \nonumber\\
&\hphantom{==}= \left|\sum_{i = 1}^{n}\EE\left[\vx{i}(1 - \vx{i})\mathbbm{1}(\varphi(X_i) - \delta - Y_i \leq F^{-1}_E(z \mid X_i)) \mid \MM\right] - z\right|\nonumber\\
&\hphantom{==}\leq \left|\sum_{i = 1}^{n}\EE\left[\vx{i}\mathbbm{1}(\varphi(X_i) - \delta - Y_i \leq F^{-1}_E(z \mid X_i)) \mid \MM\right] - z\right| + \sum_{i = 1}^{n}\EE\left[\vx{i}^2 \mid \MM\right] \nonumber\\
&\hphantom{==}\to \left|\sum_{i = 1}^{n}\EE\left[\vx{i}\mathbbm{1}(\varphi(X_i) - \delta - Y_i \leq F^{-1}_E(z \mid X_i)) \mid \MM\right] - z\right|. \label{eq:workblyh}
\end{align}

\subsection*{Proximity Conditional on Realized $\delta$-Stability}

We begin this subsection by expressing the $z$ term inside the absolute value of (\ref{eq:workblyh}) as a conditional expectation similar to the one inside the absolute value of (\ref{eq:workblyh}). In particular, we replace $z$ with the expectation given by (\ref{eq:expectedvalue}), decompose the expectation using the law of total expectation into expectations conditional on $\delta$-stability being realized or not, and eliminate leftover terms using Lemma \ref{lem:asymptotics} to obtain
\begin{align}
&\left|\sum_{i = 1}^{n}\EE[\vx{i}\mathbbm{1}(\varphi(X_i) - \delta - Y_i \leq F^{-1}_E(z \mid X_i)) \mid \MM] - z\right| \nonumber\\
&= \left|\sum_{i = 1}^{n}\EE[\vx{i}\mathbbm{1}(\varphi(X_i) - \delta - Y_i \leq F^{-1}_E(z \mid X_i)) \mid \MM] - \sum_{i = 1}^{n}\EE[\vx{i}\ux{i}]\right| \nonumber\\
&\leq \left|\sum_{i = 1}^{n}\EE[\vx{i}\mathbbm{1}(\varphi(X_i) - \delta - Y_i \leq F^{-1}_E(z \mid X_i))\mid \MM] - \sum_{i = 1}^{n}\EE[\vx{i}\ux{i} \mid \MM] \right| \nonumber\\
&\hphantom{{}\leq}+ \gamma_n\sum_{i = 1}^{n}\EE[\vx{i}\ux{i} \mid \neg\MM] + \gamma_n\sum_{i = 1}^{n}\EE[\vx{i}\ux{i} \mid \MM] \nonumber\\
&\to \left|\sum_{i = 1}^{n}\EE[\vx{i}\mathbbm{1}(\varphi(X_i) - \delta - Y_i \leq F^{-1}_E(z \mid X_i)) \mid \MM] - \sum_{i = 1}^{n}\EE[\vx{i}\ux{i} \mid \MM]\right|.\label{eq:zsub}
\end{align}
By linearity of expectation and the realized $\delta$-stability, (\ref{eq:zsub}) is bounded above by
\begin{equation}
\label{eq:workfour}
\sum_{i = 1}^{n}\EE[\vx{i}\left(\mathbbm{1}(\varphi(X_i) - \delta - Y_i \leq F^{-1}_E(z \mid X_i)) - \mathbbm{1}(\varphi(X_i) + \delta - Y_i \leq F^{-1}_E(z \mid X_i))\right) \mid \MM].
\end{equation}
Next, we show that the two indicators in (\ref{eq:workfour}) are close to each other in expectation by continuity of the CDF of $Y$ conditional on $X$. First, notice that using the tower property to condition on $X_i$ achieves independence of the indicator functions from $\vx{i}$ and obviates the conditioning on $\MM$. Therefore, (\ref{eq:workfour}) is equivalent to
\begin{align}
\sum_{i = 1}^{n}\EE[\mathbb{E}\left[\vx{i} \mid X_i\right](&\Pr(\varphi(X_i) - \delta - Y_i \leq F^{-1}_E(z \mid X_i) \mid X_i) \nonumber\\
&- \Pr(\varphi(X_i) + \delta - Y_i \leq F^{-1}_E(z \mid X_i) \mid X_i)) \mid \MM].\label{eq:almostdone}
\end{align}
By uniform continuity of the conditional CDF of $Y$ as applied in (\ref{eq:continuity}), the difference between the conditional probabilities in (\ref{eq:almostdone}) is bounded above by $\epsilon / 3$, so (\ref{eq:almostdone}) is bounded above by
\begin{equation}
\label{eq:almostdone2}
\frac{\epsilon}{3}\sum_{i = 1}^{n}\EE\left[\vx{i} \mid \MM\right].
\end{equation}
Finally, for $n$ large enough that $\gamma_n < 1/2$, $\EE\left[\vx{i} \mid \MM\right] < 2/n$ by the law of total expectation since the weights must be nonnegative and $\EE[\vx{i}] = 1/n$. Thus, (\ref{eq:almostdone2}) is bounded above by $2\epsilon / 3$. Recalling that the leftover terms we have dropped throughout these steps sum to $\epsilon / 3$, we conclude that
\[
\left|\sum_{i \neq j}\EE[\vx{i}\vx{j}\ux{i}\ux{j}] - z^2\right| < \epsilon,
\]
as desired.
\end{proof}

\begin{lemma2}
\label{lem:shifttermsumofvar}
Under Assumptions \ref{as:uniform}-\ref{as:weight},
\[
\sum_{i = 1}^{n}\Var\left(\vx{i}\left(\mathbbm{1}(U_i \leq F_E(e \mid X_i)) - F_E(e \mid X_i)\right)\right) \to 0,\hspace{20pt}n \to \infty.
\]
\end{lemma2}
\begin{proof}[Proof of Lemma \ref{lem:shifttermsumofvar}]
We decompose the sum of variances in (\ref{eq:part2vardecompose}) via the law of total variance:
\begin{align}
&\sum_{i = 1}^{n}\Var\left(\vx{i}\left(\mathbbm{1}(U_i \leq F_E(e \mid X_i)) - F_E(e \mid X_i)\right)\right) \nonumber\\
&\hphantom{====}= \sum_{i = 1}^{n}\Var\left(\mathbb{E}\left[\vx{i}\left(\mathbbm{1}(U_i \leq F_E(e \mid X_i)) - F_E(e \mid X_i)\right) \mid \Omega_v \setminus \{Y_i\}\right]\right) \nonumber\\
&\hphantom{===={}=\sum_{i = 1}^{n}}+ \mathbb{E}\left[\Var\left(\vx{i}\left(\mathbbm{1}(U_i \leq F_E(e \mid X_i)) - F_E(e \mid X_i)\right) \mid \Omega_v \setminus \{Y_i\}\right)\right]. \label{eq:totvar}
\end{align}
By noting that $\vx{i}$ is a constant given $\Omega_v \setminus \{Y_i\}$ and applying an argument similar to the one yielding (\ref{eq:tower1}), we can reduce the variance-of-expectation term in (\ref{eq:totvar}) to
\begin{align*}
\sum_{i = 1}^{n}&\Var\left(\mathbb{E}\left[\vx{i}\left(\mathbbm{1}(U_i \leq F_E(e \mid X_i)) - F_E(e \mid X_i)\right) \mid \Omega_v \setminus \{Y_i\}\right]\right)\\
&= \sum_{i = 1}^{n}\Var\left(\vx{i}\mathbb{E}\left[\left(\mathbbm{1}(U_i \leq F_E(e \mid X_i)) - F_E(e \mid X_i)\right) \mid \Omega_v \setminus \{Y_i\}\right]\right)\\
&= \sum_{i = 1}^{n}\Var\left(\vx{i}\mathbb{E}\left[\left(\mathbbm{1}(U_i \leq F_E(e \mid X_i)) - F_E(e \mid X_i)\right) \mid X_i\right]\right)\\
&= \sum_{i = 1}^{n}\Var\left(\vx{i}\left(\Pr(U_i \leq F_E(e \mid X_i)\mid X_i) - \mathbb{E}\left[F_E(e \mid X_i) \mid X_i\right]\right)\right)\\
&= 0.
\end{align*}
Moreover, we can reduce the expectation-of-variance term in (\ref{eq:totvar}) using Assumption \ref{as:propobs} to note that the maximum possible weight $M_n$ of an observation converges to zero in $n$:
\begin{align}
\sum_{i = 1}^{n}\mathbb{E}&\left[\Var\left(\vx{i}\left(\mathbbm{1}(U_i \leq F_E(e \mid X_i)) - F_E(e \mid X_i)\right) \mid \Omega_v \setminus \{Y_i\}\right)\right] \nonumber\\
&= \sum_{i = 1}^{n}\mathbb{E}\left[\vx{i}^2\Var\left(\mathbbm{1}(U_i \leq F_E(e \mid X_i)) - F_E(e \mid X_i) \mid X_i\right)\right] \nonumber\\
&\lesssim\sum_{i = 1}^{n}\mathbb{E}\left[\vx{i}^2\right] \nonumber\\
&\leq M_n\sum_{i = 1}^{n}\mathbb{E}\left[\vx{i}\right] \nonumber\\
&= M_n \nonumber\\
&\to 0. \nonumber
\end{align}
This completes the proof.
\end{proof}

\begin{lemma2}
\label{lem:shifttermsumofcov}
Under Assumptions \ref{as:uniform}-\ref{as:weight},
\begin{align*}
\sum_{i \neq j}\text{Cov}&\Big(\vx{i}\big(\mathbbm{1}(U_i \leq F_E(e \mid X_i)) - F_E(e \mid X_i)\big), \vx{j}\big(\mathbbm{1}(U_j \leq F_E(e \mid X_j)) - F_E(e \mid X_j)\big)\Big)\\
&\to 0
\end{align*}
as $n \to \infty$.
\end{lemma2}
\begin{proof}[Proof of Lemma \ref{lem:shifttermsumofcov}]
First, we rewrite the covariance in terms of expectations:
\begin{align*}
&\sum_{i \neq j}\text{Cov}\left(\vx{i}(\mathbbm{1}(U_i \leq F_E(e \mid X_i)) - F_E(e \mid X_i)), \vx{j}(\mathbbm{1}(U_j \leq F_E(e \mid X_j)) - F_E(e \mid X_j))\right)\\
&= \sum_{i \neq j}\mathbb{E}\left[\vx{i}\vx{j}(\mathbbm{1}(U_i \leq F_E(e \mid X_i)) - F_E(e \mid X_i))(\mathbbm{1}(U_j \leq F_E(e \mid X_j)) - F_E(e \mid X_j))\right]\\
&\hphantom{{}=}- \mathbb{E}\left[\vx{i}(\mathbbm{1}(U_i \leq F_E(e \mid X_i)) - F_E(e \mid X_i))\right]\mathbb{E}\left[\vx{j}(\mathbbm{1}(U_j \leq F_E(e \mid X_j)) - F_E(e \mid X_j))\right]\\
&= \sum_{i \neq j}\mathbb{E}\left[\vx{i}\vx{j}(\mathbbm{1}(U_i \leq F_E(e \mid X_i)) - F_E(e \mid X_i))(\mathbbm{1}(U_j \leq F_E(e \mid X_j)) - F_E(e \mid X_j))\right],
\end{align*}
where the last equality follows by (\ref{eq:tower1}). We therefore seek to show that
\[
\sum_{i \neq j}\mathbb{E}\left[\vx{i}\vx{j}(\mathbbm{1}(U_i \leq F_E(e \mid X_i)) - F_E(e \mid X_i))(\mathbbm{1}(U_j \leq F_E(e \mid X_j)) - F_E(e \mid X_j))\right] \to 0.
\]
Let $\epsilon > 0$. As before, the uniform continuity of the conditional response distribution implies that there exists $\delta_1 > 0$ so that
\begin{equation}
\label{eq:continuity2}
\left|y_1 - y_2\right| < 2\delta_1 \implies \left|F_{Y}(y_1 \mid x) - F_{Y}(y_2 \mid x)\right| < \epsilon / 3.
\end{equation}
Fix $\delta < \min\left\{\delta_0, \delta_1\right\}$. Then, by Lemma \ref{lem:asymptotics}, there exists $N \in \mathbb{N}$ so that, for $n \geq N$, terms identified in Lemma \ref{lem:asymptotics} that appear in this proof sum to at most $\epsilon / 3$; for concision, we note these leftover terms where they appear and then cite Lemma \ref{lem:asymptotics} to drop them.

\subsection*{Conditioning on and Applying Realized $\delta$-Stability}

We condition on the event that $\delta$-stability is realized using the law of total expectation, then apply the triangle inequality and Lemma \ref{lem:asymptotics} to bound
\[
\left|\sum_{i \neq j}\mathbb{E}\left[\vx{i}\vx{j}(\mathbbm{1}(U_i \leq F_E(e \mid X_i)) - F_E(e \mid X_i))(\mathbbm{1}(U_j \leq F_E(e \mid X_j)) - F_E(e \mid X_j))\right]\right|
\]
above by
\begin{equation}
\label{eq:haha}
\left|\sum_{i \neq j}\mathbb{E}\left[\vx{i}\vx{j}(\mathbbm{1}(U_i \leq F_E(e \mid X_i)) - F_E(e \mid X_i))(\mathbbm{1}(U_j \leq F_E(e \mid X_j)) - F_E(e \mid X_j)) \mid \MM\right]\right|,
\end{equation}
discounting leftover terms. We then expand (\ref{eq:haha}) to
\begin{align}
&\left|\sum_{i \neq j}\mathbb{E}\left[\vx{i}\vx{j}\mathbbm{1}(U_i \leq F_E(e \mid X_i))(\mathbbm{1}(U_j \leq F_E(e \mid X_j)) - F_E(e \mid X_j)) \mid \MM\right]\right. \nonumber\\
&\hphantom{\Bigg|}\left. - \sum_{i \neq j}\mathbb{E}\left[\vx{i}\vx{j}F_E(e \mid X_i)(\mathbbm{1}(U_j \leq F_E(e \mid X_j)) - F_E(e \mid X_j)) \mid \MM\right]\right|. \label{eq:blah47}
\end{align}
We then apply the realized $\delta$-stability of $\hat{\varphi}(X_i)$ to the first summation in (\ref{eq:blah47}) to show that
\begin{align}
&\left|\sum_{i \neq j}\mathbb{E}\left[\vx{i}\vx{j}\mathbbm{1}(U_i \leq F_E(e \mid X_i))(\mathbbm{1}(U_j \leq F_E(e \mid X_j)) - F_E(e \mid X_j)) \mid \MM\right]\right. \nonumber\\
&\hphantom{\Bigg|}\left.- \sum_{i \neq j}\mathbb{E}\left[\vx{i}\vx{j}\mathbbm{1}(\varphi(X_i) - Y_i + \delta \leq e)(\mathbbm{1}(U_j \leq F_E(e \mid X_j)) - F_E(e \mid X_j)) \mid \MM\right]\right| < \frac{2\epsilon}{3}. \label{eq:applystable}
\end{align}
Our proof of this claim is as follows. We can collapse terms inside the absolute value of (\ref{eq:applystable}) to
\begin{equation}
\label{eq:474747}
\left|\sum_{i \neq j}\mathbb{E}\left[\vx{i}\vx{j}(\mathbbm{1}(\hat{\varphi}(X_i) - Y_i \leq e) - \mathbbm{1}(\varphi(X_i) - Y_i + \delta \leq e))(\mathbbm{1}(U_j \leq F_E(e \mid X_j)) - F_E(e \mid X_j)) \mid \MM\right]\right|.
\end{equation}
Next, because, conditional on $\MM$, $\mathbbm{1}(\hat{\varphi}(X_i) - Y_i \leq e) - \mathbbm{1}(\varphi(X_i) - Y_i + \delta \leq e) \geq 0$, we can bound (\ref{eq:474747}) above via the triangle inequality and Jensen's inequality by
\begin{equation}
\label{eq:47474747}
\sum_{i = 1}^{n}\mathbb{E}\left[\vx{i}(\mathbbm{1}(\hat{\varphi}(X_i) - Y_i \leq e) - \mathbbm{1}(\varphi(X_i) - Y_i + \delta \leq e)) \mid \MM\right].
\end{equation}
Note that, conditional on $\MM$, $\mathbbm{1}(\varphi(X_i) - Y_i - \delta \leq e) \geq \mathbbm{1}(\hat{\varphi}(X_i) - Y_i \leq e)$. Using this fact and applying the tower property to condition on $X_i$, we can bound (\ref{eq:47474747}) above by
\begin{equation}
\label{eq:asdf123}
\sum_{i = 1}^{n}\mathbb{E}\left[\EE\left[\vx{i} \mid X_i\right]\left(\Pr(\varphi(X_i) - Y_i - \delta \leq e \mid X_i) - \Pr(\varphi(X_i) - Y_i + \delta \leq e \mid X_i)\right) \mid \MM\right].
\end{equation}
By uniform continuity of the conditional distribution of $Y$ given $X$ as applied in (\ref{eq:continuity2}) and the fact that $\EE\left[\vx{i} \mid \MM\right] < 2 / n$ for $n$ large enough that $\gamma_n < 1 / 2$, (\ref{eq:asdf123}) can be bounded above by
\[
\frac{\epsilon}{3}\sum_{i = 1}^{n}\EE\left[\vx{i} \mid \MM\right] < \frac{\epsilon}{3}\sum_{i = 1}^{n}\frac{2}{n} = \frac{2\epsilon}{3}.
\]
Thus, we have shown (\ref{eq:applystable}). Applying this result to (\ref{eq:blah47}) via the triangle inequality and re-collapsing terms, we can bound (\ref{eq:blah47}) above by
\begin{equation}
\label{eq:blah4747}
\left|\sum_{i \neq j}\mathbb{E}\left[\vx{i}\vx{j}(\mathbbm{1}(\varphi(X_i) - Y_i + \delta \leq e) - F_E(e \mid X_i))(\mathbbm{1}(U_j \leq F_E(e \mid X_j)) - F_E(e \mid X_j)) \mid \MM\right]\right| + \frac{2\epsilon}{3}.
\end{equation}

\subsection*{Conclusion}

Lastly, we use the law of total expectation to decompose the absolute value term in (\ref{eq:blah4747}) as
\begin{align}
&\left|\sum_{i \neq j}\mathbb{E}\left[\vx{i}\vx{j}(\mathbbm{1}(\varphi(X_i) - Y_i + \delta \leq e) - F_E(e \mid X_i))(\mathbbm{1}(U_j \leq F_E(e \mid X_j)) - F_E(e \mid X_j)) \mid \MM\right]\right| \nonumber\\
&= \left|\frac{1}{1 - \gamma_n}\sum_{i \neq j}\left(\underbrace{\mathbb{E}\left[\vx{i}\vx{j}(\mathbbm{1}(\varphi(X_i) - Y_i + \delta \leq e) - F_E(e \mid X_i))(\mathbbm{1}(U_j \leq F_E(e \mid X_j)) - F_E(e \mid X_j))\right]}_{= 0\text{ by the tower property conditioning on }X_j}\right.\right. \nonumber\\
&\left.\left.\vphantom{\sum_{i \neq j}}\hphantom{\Bigg|}- \gamma_n\mathbb{E}\left[\vx{i}\vx{j}(\mathbbm{1}(\varphi(X_i) - Y_i + \delta \leq e) - F_E(e \mid X_i))(\mathbbm{1}(U_j \leq F_E(e \mid X_j)) - F_E(e \mid X_j)) \mid \neg\MM\right]\right)\right| \nonumber\\
&\leq \frac{\gamma_n}{1 - \gamma_n}\sum_{i \neq j}\mathbb{E}\left[\vx{i} \mid \neg\MM\right] \label{eq:hah}\\
&\to 0, \nonumber
\end{align}
where (\ref{eq:hah}) follows by the triangle inequality and Jensen's inequality. We thus have that (\ref{eq:blah4747}) is bounded above by $2\epsilon / 3$. Recalling that the leftover terms we have dropped throughout these steps sum to $\epsilon / 3$, we conclude that
\[
\left|\sum_{i \neq j}\mathbb{E}\left[\vx{i}\vx{j}(\mathbbm{1}(U_i \leq F_E(e \mid X_i)) - F_E(e \mid X_i))(\mathbbm{1}(U_j \leq F_E(e \mid X_j)) - F_E(e \mid X_j))\right]\right| < \epsilon,
\]
which completes the proof.
\end{proof}

\newpage

\bibliography{bibliography}

\end{document}